\documentclass[conference]{IEEEtran}
\usepackage{times}

\usepackage[numbers]{natbib}
\usepackage{multicol}
\usepackage[bookmarks=true]{hyperref}
\usepackage{algorithm}
\usepackage{algpseudocode}
\usepackage[dvipsnames]{xcolor}
\usepackage{graphicx} % DO NOT CHANGE THIS
\usepackage{mathtools}
\usepackage{amsmath}
\usepackage{amssymb}
\usepackage{url}
\usepackage{cleveref}
\usepackage{pdfpages}
\usepackage{dblfloatfix} % Add this in your preamble

\let\labelindent\relax
\usepackage{enumitem}% These are are recommended to typeset listings but not required. See the subsubsection on listing. Remove this block if you don't have listings in your paper.
\usepackage{newfloat}
\usepackage{listings}
\usepackage{wrapfig}
\usepackage{silence}
\newcommand{\xiaomin}[1]{{\color{black} #1 \color{black}}}

\newcommand{\yantian}[1]{{\color{black} #1 \color{black}}}
\def\mund#1{\noindent{\bf #1}:}
\def\qund#1{\noindent{\bf #1}}

\begin{document}

\title{\LARGE \bf
\textcolor{Aquamarine}{\huge{AAM-SEALS}}: Developing \textcolor{Aquamarine}{\huge{A}}erial-\textcolor{Aquamarine}{\huge{A}}quatic \textcolor{Aquamarine}{\huge{M}}anipulators \\in \textcolor{Aquamarine}{\huge{SE}}a, \textcolor{Aquamarine}{\huge{A}}ir, and \textcolor{Aquamarine}{\huge{L}}and \textcolor{Aquamarine}{\huge{S}}imulator
}

\author{
  {William Yang$^*$, Karthikeya Kona$^*$, Yashveer Jain$^*$, Tomer Atzili$^*$, Abhinav Bhamidipati, Xiaomin Lin, Yantian Zha}\\
  % Department of Electrical Engineering and Computer Sciences\\
  University of Maryland, College Park\\
  % United States\\
  \texttt{\{wyang124,kkona,yashveer,tatzili,abhinav7,xlin01,ytzha\}@umd.edu} \\
  %% examples of more authors
  %% \And
  %% Coauthor \\
  %% Affiliation \\
  %% Address \\
  %% \texttt{email} \\
  %% \AND
  %% Coauthor \\
  %% Affiliation \\
  %% Address \\
  %% \texttt{email} \\
  %% \And
  %% Coauthor \\
  %% Affiliation \\
  %% Address \\
  %% \texttt{email} \\
  %% \And
  %% Coauthor \\
  %% Affiliation \\
  %% Address \\
  %% \texttt{email} \\
}

%\author{\authorblockN{Michael Shell}
%\authorblockA{School of Electrical and\\Computer Engineering\\
%Georgia Institute of Technology\\
%Atlanta, Georgia 30332--0250\\
%Email: mshell@ece.gatech.edu}
%\and
%\authorblockA{Twentieth Century Fox\\
%\and
%\authorblockN{James Kirk\\ and Montgomery Scott}
%\authorblockA{Starfleet Academy\\
%San Francisco, California 96678-2391\\

% avoiding spaces at the end of the author lines is not a problem with
% conference papers because we don't use \thanks or \IEEEmembership

%Homer Simpson\authorrefmark{2},
%Montgomery Scott\authorrefmark{3} and
%Eldon Tyrell\authorrefmark{4}}
%Telephone: (800) 555--1212, Fax: (888) 555--1212}
%\authorblockA{\authorrefmark{4}Tyrell Inc., 123 Replicant Street, Los Angeles, California 90210--4321}}

\makeatletter
\let\@oldmaketitle\@maketitle% Store \@maketitle
\renewcommand{\@maketitle}{%
  \@oldmaketitle% Call the original \@maketitle
  \begin{center}
    \refstepcounter{figure} % Increment the figure counter
    \includegraphics[width=\linewidth]{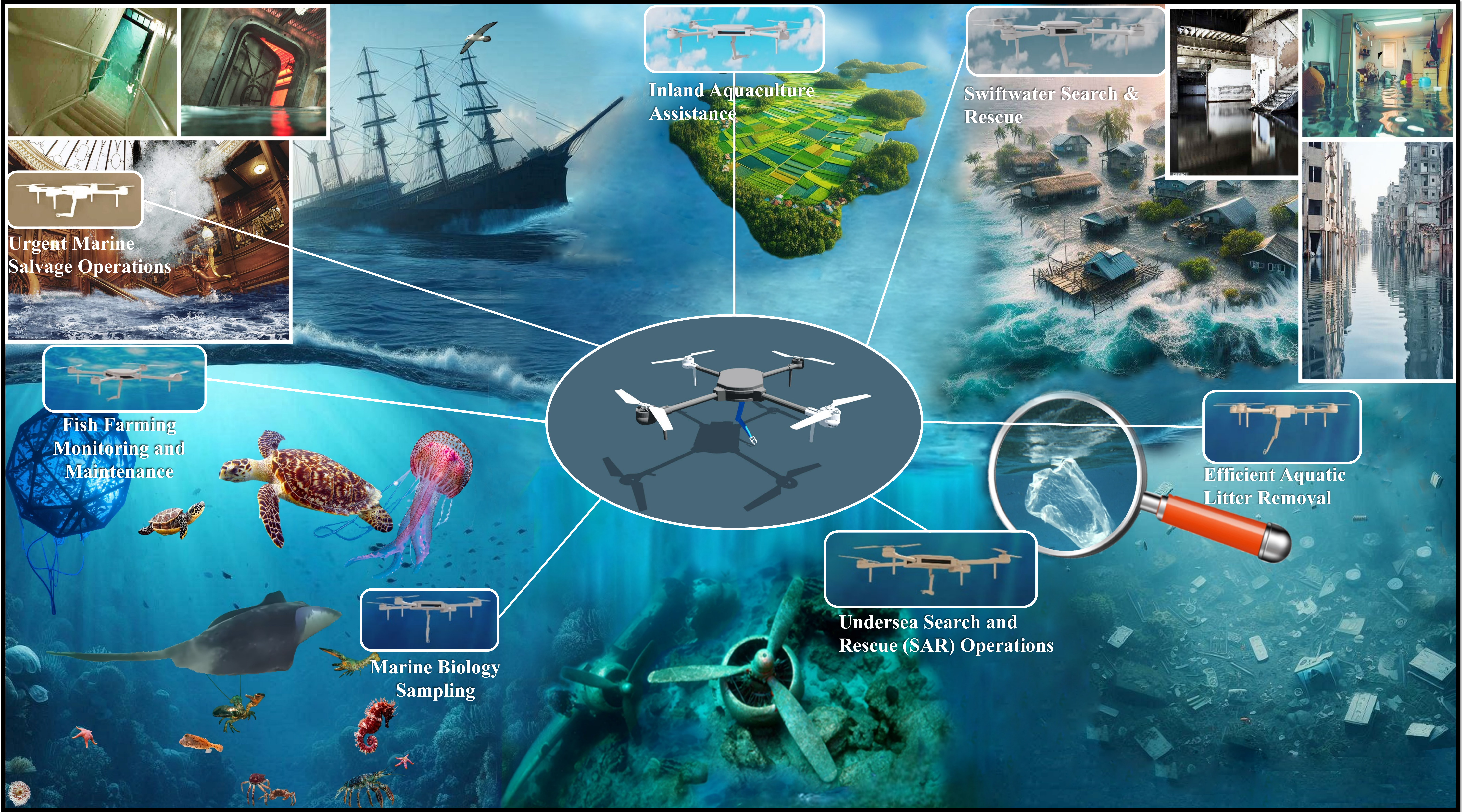}%
    \par\smallskip
    Fig. 1. Showing a wide range of critical applications that leverage AAMs' unique capabilities across sea, air, and land.
    \label{fig:intro} % Define a label for referencing
  \end{center}
  \bigskip
  \setcounter{figure}{1}\vspace*{-5mm} % Reset figure counter to 0 after including the title figure
}
\makeatother

\maketitle
\def\thefootnote{*}\footnotetext{The first four authors contributed equally, and are listed in reverse alphabetical order.}\def\thefootnote{\arabic{footnote}}

\begin{abstract}
Current mobile manipulators and high-fidelity simulators lack the ability to seamlessly operate and simulate across integrated environments spanning sea, air, and land. To address this gap, we introduce Aerial-Aquatic Manipulators (AAMs) in SEa, Air, and Land Simulator (SEALS), a comprehensive and photorealistic simulator designed for AAMs to operate and learn in these diverse environments. The development of AAM-SEALS tackles several significant challenges, including the creation of integrated controllers for flying, swimming, and manipulation, and the high-fidelity simulation of aerial dynamics and hydrodynamics leveraging particle-based hydrodynamics. Our evaluation demonstrates smooth operation and photorealistic transitions across air, water, and their interfaces. We quantitatively validate the fidelity of particle-based hydrodynamics by comparing position-tracking errors across real-world and simulated systems. AAM-SEALS benefits a broad range of robotics communities, including robot learning, aerial robotics, underwater robotics, mobile manipulation, and robotic simulators. We will open-source our code and data to foster the advancement of research in these fields. The overview video is available at \url{https://youtu.be/MbqIIrYvR78}. Visit our project website at \url{aam-seals.umd.edu} for more details. 
\end{abstract}

\IEEEpeerreviewmaketitle

% \vspace*{-17mm}
\section{INTRODUCTION}

Mobile manipulation is a crucial and rapidly advancing field in robotics, offering the potential to revolutionize various industries by enabling robots to interact with and manipulate their environments. This capability is especially valuable in scenarios that are tedious, hazardous, or challenging for humans. Despite its significance, current research has focused predominantly on mobile manipulation in isolated environments -- either in the sea, air, or on land. For instance, aerial manipulation involves robots performing tasks while flying, underwater manipulation focuses on submersible robots operating in aquatic environments, and ground-based mobile manipulation deals with robots navigating and interacting on terrestrial surfaces.

However, many real-world applications require robots to operate seamlessly across different environments. For instance, an ideal robotic system for search and rescue missions might need to take off from the ground and navigate through the air for most of the journey, before diving into water to reach and assist victims efficiently. This necessitates mobile manipulators capable of transitioning and functioning effectively across water, air, and land boundaries, as illustrated in Fig.~\ref{fig:intro}.

To address this need, we propose a novel class of robots called Aerial-Aquatic Manipulators (AAMs). AAMs combine the capabilities of aerial manipulators \cite{khamseh2018aerial,alzu2018loon,abdulmajeed2019autonomous,samadikhoshkho2020nonlinear,imanberdiyev2019fast}, underwater manipulators \cite{lee2000robust,zhang2015design,youakim2018motion,sivvcev2018underwater,cieslak2015autonomous}, and aerial-aquatic quadrotors \cite{tan2019design,tan2020morphable,wu2024design,semenov2023control,liu2023tj}. Our AAMs have unique advantages, such as the ability to navigate large areas efficiently and adaptively select the safest or most efficient path. For instance, an AAM can fly out of a debris-filled water area, travel through the air to a new location, and then re-enter the water to reach a target area.

The design and construction of such advanced robots poses significant challenges. Directly developing a physical AAM is complex and expensive, involving intricate designs for sensors, mechanics, morphologies, kinematics, and robot-environment interactions. To mitigate these risks and costs, we first propose developing AAMs within a high-fidelity simulation environment. This approach allows us to validate our designs and refine them iteratively, enabling testing, evaluation, validation, and verification (TEVV) before physical implementation.

In this paper, we introduce AAM-SEALS, a comprehensive and photorealistic simulator built on top of NVIDIA Isaac Sim \cite{liang2018gpu}. AAM-SEALS enables Aerial-Aquatic Manipulators (AAMs) to operate and learn in integrated environments that encompass sea, air, and land (SEALS). Developing AAM-SEALS involved addressing several significant challenges, including the creation of integrated controllers for both flight and manipulation, as well as the high-fidelity simulation of aerial dynamics and hydrodynamics using particle physics \cite{fraga2019smoothed,macklin2013position,macklin2014unified,muller2007position}. Particle-based hydrodynamics directly models fluids as a set of particles that interact with and constrain each other and surrounding objects, enabling the simulation of complex fluid dynamics and object-fluid interactions. This approach aligns with our goal of enabling AAMs to operate effectively in free-surface flows \cite{king1987free,violeau2016smoothed}.

% We comprehensively evaluate our AAM-SEALS system across various aspects. Firstly, we showcase the teleoperation prowess of our AAM within the SEALS environment. Secondly, we conduct a quantitative comparison between the hydrodynamics of our SEALS, employing particle physics, and the conventional rigid body hydrodynamics used in the cutting-edge photorealistic underwater simulator, UNav-Sim \cite{amer2023unav}. This comparison entails analyzing position tracking and discerning photo-realism disparities, thus demonstrating the high-fidelity and photorealistic fluid simulations achievable with SEALS. Third, we demonstrate the potential of using AAM-SEALS for robot learning by conducting visual reinforcement learning experiments. %Third, we demonstration the our AAM is functional in SEALS via teleoperation, and quantitively compare the convergence of position-tracking error in particle-based and rigid-body-based hydrodynamics.

%Third, we evaluate state-of-the-art imitation learning and reinforcement learning techniques on realistically valuable tasks, such as catching crabs and long-horizon missions, within the AAM-SEALS framework.
\yantian{
Our contributions are threefold, spanning robot design, simulation development, and application demonstration: 

\noindent\textbf{Robot:} We introduce AAMs, a novel class of robots designed for versatile cross-medium tasks. We developed an exemplary model of AAMs, showcasing novel morphology, dynamic capabilities, and a sophisticated control system that can adapt to changing centers of gravity. The control system's effectiveness was assessed using position-tracking error, a widely-adopted metric for evaluating control systems.

\noindent\textbf{Simulation:} We created SEALS, a photorealistic, high-fidelity simulation environment for AAMs, and thoroughly evaluated the fidelity of both air and water dynamics through comparisons with real-world and simulated environments. 

\noindent\textbf{Application:} We introduced a novel aerial-aquatic manipulation challenge centered on searching for and capturing moving aquatic animals, such as crabs. To demonstrate the effectiveness of our AAM and SEALS, we developed a unique dataset featuring fully controllable aquatic animal models and specialized tools for collecting demonstration trajectories for AAM-based animal capture. Additionally, we implemented state-of-the-art reinforcement learning from demonstrations (RLfD) and visual reinforcement learning to facilitate future research in aerial-aquatic manipulation.
}
% Our contributions are threefold. Robot: 1) we introduce a novel class of robots, Aerial-Aquatic Manipulators (AAMs), capable of performing a wide range of highly valuable cross-medium tasks; 2) we developed a representative exmple of morphology, dynamics, and advanced control that can handel changing center of gravity. Simulator: 1) we develop a photorealistic, high-fidelity simulation environment tailored specifically for AAMs; 2) we throughly evaluated the fidelity of air and water in SEALS by conducting both real world and simulated environments. Application: we introduce a novel cross-medium manipulation task of searching and catching moving animals such as crabs; we demonstrate the effectiveness of our AAM and SEALS by successfully teleoperating our AAM to perform such challenging tasks, and leading to a AAM crab catching dataset including 40 demonstration trajectories; 3) we conduct cutting-edge visual reinforcement learning. 

%\yantian{This work, encompassing both novel robotic systems and advanced simulation tools, aligns with key ICLR themes such as \textit{applications to robotics and autonomy} as well as \textit{infrastructure, software libraries, etc.} Additionally, the recent acceptance of multiple robotics simulators at ICLR, including Meta-Habitat 3.0\footnote{\url{https://openreview.net/forum?id=4znwzG92CE}}, signals a growing interest in this area within the ICLR community.}

\section{RELATED WORK} \label{sec:related_work}

\mund{Aerial-Aquatic Quadrotors}
The development of hybrid aerial-aquatic quadrotors has recently gained significant interest due to the popularity of quadrotors and the broad needs of tasks such as filming and aerial-aquatic environmental monitoring. Tan and Chen \cite{tan2020morphable} developed a morphable aerial-aquatic quadrotor with symmetric thrust vectoring to adapt thrust direction for optimal performance in both air and water. They further explored this concept by integrating multi-rotors to refine propulsion systems and mechanical design \cite{tan2019design}. Alzu'bi et al. \cite{alzu2018loon} introduced the Loon Copter, a hybrid vehicle with active buoyancy control for smooth transitions between air and water, suitable for underwater exploration and environmental monitoring. Wu et al. \cite{wu2024design} demonstrated a tandem dual rotor aerial-aquatic vehicle focusing on efficient propulsion and maneuverability. Liu et al. \cite{liu2023tj} advanced the field with the TJ-FlyingFish, which features tilt-able propulsion units for improved stability and control in both environments. These works collectively highlight significant progress in hybrid aerial-aquatic vehicles, showcasing innovative approaches to overcome the unique challenges of operating in both air and water. However, they have not considered the addition of manipulators which would drastically enlarge the number of tasks, and effective simulation tools are essential for the further development and testing of these hybrid systems. 

\mund{Photorealistic Aerial or Underwater Simulators}
Simulation environments are crucial for both gathering data and fostering the acquisition of new capabilities by robots. Advanced aerial robotics simulators such as Pegasus \cite{jacinto2023pegasus}, built on IsaacSim \cite{liang2018gpu}, and AirSim \cite{shah2018airsim} provide high-fidelity rendering. However, simulating underwater environments presents greater challenges. Recent advances have targeted complex underwater environments and maritime scenarios. For example, Zwilgmeyer et al. \cite{zwilgmeyer2021creating} use Blender to generate underwater datasets, while platforms such as UUV Simulator \cite{manhaes2016uuv} and UWSim \cite{dhurandher2008uwsim} model underwater physics and sensors. Despite their progress, these efforts have been discontinued. DAVE \cite{zhang2022dave} seeks to bridge this gap but struggles with rendering limitations.

More recent simulators such as HoloOcean \cite{potokar2022holoocean}, MARUS \cite{lonvcar2022marus}, and UNav-Sim \cite{amer2023unav} have improved rendering realism but still struggle to simulate complex free-space fluids and object-water interactions without using particle physics. AuqaSim \cite{wu2024marvis} focuses on near-water tasks, but lacks drone simulation above the water. Many simulators built on Unreal Engine face modifiability challenges and often do not release their original project files. ChatSim \cite{palnitkar2023chatsim} integrates ChatGPT with OysterSim \cite{lin2022oystersim}, enabling easy modifications of the simulated environment and generating photorealistic underwater settings. However, these simulators mainly address deep underwater tasks and often neglect aerial parts and air-water transitions.

\section{\textcolor{Aquamarine}{\textbf{A}}erial-\textcolor{Aquamarine}{\textbf{A}}quatic \textcolor{Aquamarine}{\textbf{M}}anipulator (\textcolor{Aquamarine}{\textbf{AAM}})}
\subsection{AAM Dynamics Modeling} \label{sssec:AAMDM}

\yantian{
Aerial-Aquatic Manipulation is a novel concept introduced in this work, promising to open a new field of research. The general modeling of Aerial Aquatic Manipulators (AAMs) involves a cross-medium drone platform with $n$ thrusters $T1...Tn$, a manipulator with $m$ degrees of freedom (DoF), and a multi-finger gripper. For simplicity, we use a representative example of an AAM consisting of an aerial-aquatic quadrotor with four thrusters, a manipulator with three DoF, and a three-finger gripper. The quadrotor is chosen for its widespread use and robust performance in tasks such as object retrieval and handling. The manipulator’s design enables precise and versatile operations, making it adaptable to a range of environments and tasks.

This AAM model is versatile and can be adapted to other aerial-aquatic platforms, such as hexacopters, or systems with different manipulator configurations and grippers, with minimal modifications. Our AAM serves as a prime example, and the simulator SEALS (discussed in Sec.~\ref{sec:SEALS}) has been developed to allow researchers to test and refine various AAM designs with reduced costs and risks. We also included a guideline for future researchers to create their customized AAMs in Appendix.~\ref{Ap:custom_aam}. 
}

\vspace*{-5mm}
\begin{figure}[H]
  \begin{center}
    \includegraphics[width=0.275\textwidth]{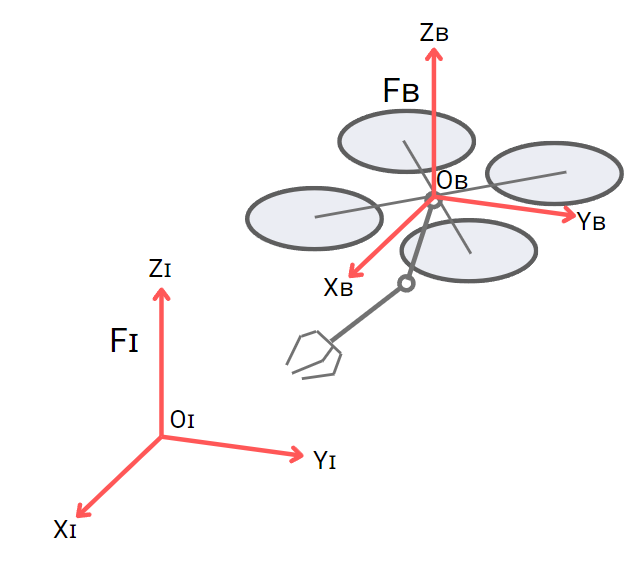}
  \end{center}
  \label{fig:sch_view}
  \vspace{-6mm}\caption{Schematic of a representative Aerial Aquatic Manipulator.}
  \vspace{-2mm}
\end{figure}

The AAM's simulation follows the conventions outlined in the Isaac Sim simulator. Isaac Sim employs a right-handed rule convention where the Z-axis of the inertial frame points upwards, and the Y-axis is aligned with true North, adhering to the East-North-Up (ENU) coordinate system. For the vehicle's body frame, a front-left-up (FLU) convention is adopted. \cite{jacinto2023pegasus} This standardized coordinate system facilitates the integration and simulation of AAM's movements and operations within the virtual environment, ensuring consistency and accuracy in control and navigation algorithms.

% \begin{figure}
%     \centering
%     \includegraphics[width=0.35\linewidth]{figures/ref frame.png}
%     \caption{Schematic View of AAM}
%     \label{fig:schematic_view}
% \end{figure}

In Fig.~\ref{fig:sch_view}., the coordinate frame of inertial is denoted with $C_I:\{ X_I, Y_I, Z_I \}$ with the Origin $O_I$, while the coordinate frame of the body is denoted with $C_B : \{ X_B, Y_B, Z_B \}$ with the Origin $O_B$ indicating the center of mass of AAM. 

The angular velocity can be expressed as: 
\begin{equation}
    \dot \omega = J^{-1} (\tau - \omega \times J\omega )
\end{equation}
where, $J$ is the inertia tensor for vehicle expressed in $C_B$, $\omega $ denotes the angular velocity of $C_B$ with respect to $C_I$ expressed in $C_B$, $\tau$ denotes the total torque from each rotor. \yantian{Following a recent work by Jacinto et al. (2023) \cite{jacinto2023pegasus}, we can compute $\tau$ by multiplying the forces of individual rotors, represented as the vector $\mathbf{F} = [F_1, \dots, F_N]$, with an allocation matrix $\mathbf{A}$:

\begin{equation} \label{eq:force_relation}
    \tau = \mathbf{A} \mathbf{F}
\end{equation} where the allocation matrix $\mathbf{A}$ is computed based on the quadrotor parameters including the arm length and rotor positions. We will define $\mathbf{A}$ in the following section.

However, most of the work on drone control assumes a fixed center of gravity (CoG), which cannot satisfy our needs of aerial manipulation. We now explain our improvements to handle dynamic changes of CoG in the next subsection, Sec.~\ref{sec:CoG}. 

The dynamics of a 3-DoF manipulator is shown in Eq.~\ref{eq:manipulator_kine} (essentially a kinematic equation).
The kinematic equation connects how the joints move (joint velocities $\dot{q}$) to how the end of the robot moves (end-effector velocities $\dot{x}$). It uses a Jacobian matrix ($\mathbf{J}(q)$) to calculate these velocities. The manipulator motion is determined using inverse kinematics, allowing the calculation of the joint velocities $\dot{q}$ required to achieve the desired end-effector velocity $\dot{x}$. 
\begin{equation} \label{eq:manipulator_kine}
    \dot{x} = \mathbf{J}(q) \dot{q}    
\end{equation}
where $\dot{x}$ is the end-effector velocity, $\dot{q}$ is the joint velocity, and $\mathbf{J}(q)$ is the Jacobian matrix.

\begin{figure*}[t]
    \centering
    \includegraphics[width=1\linewidth]{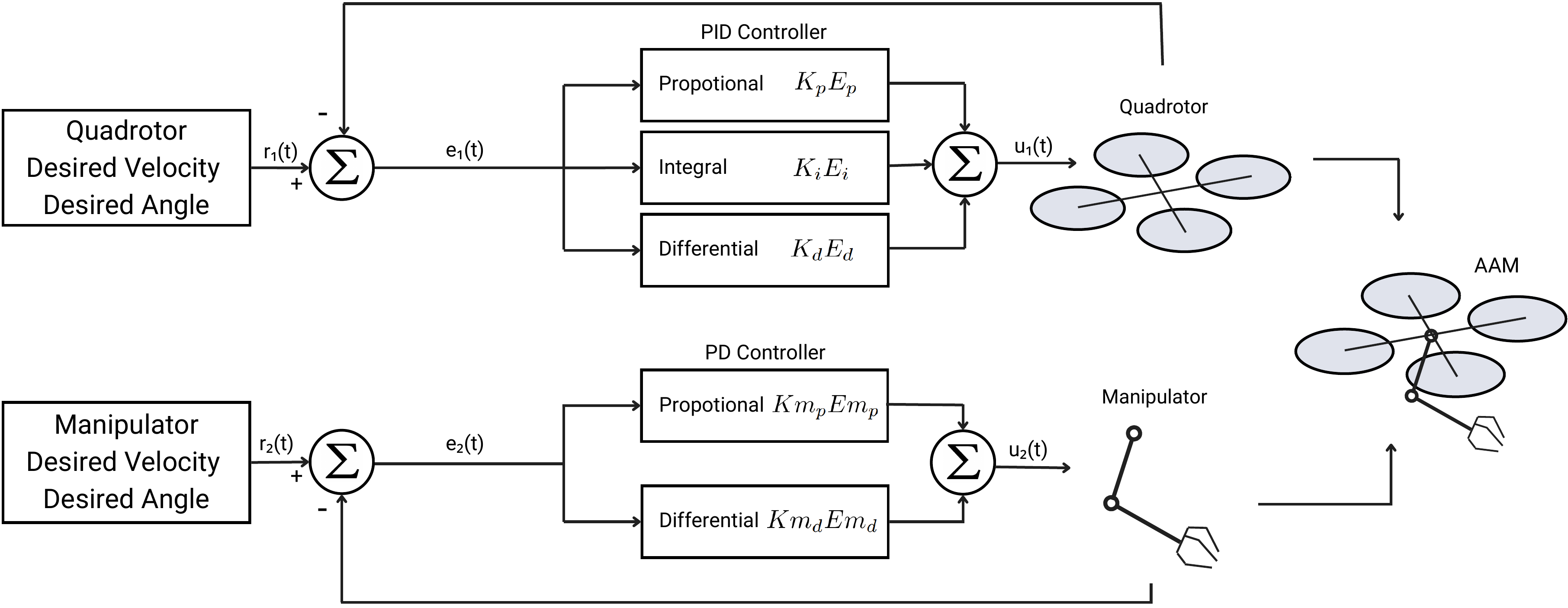}
    \caption{\yantian{Block diagram of controller for AAM}}
    \label{fig:block_diagram}
\end{figure*}

\subsection{Handling the Change of Center of Gravity (CoG)} \label{sec:CoG}
In our simulator, we handle dynamic changes in the center of gravity (CoG) of the quadcopter by continuously updating the allocation matrix at each time step. Therefore, our unique design of the allocation matrix $\mathbf{A}$ becomes:

% \begin{equation}
% \mathbf{A} =
% \begin{bmatrix}
% k_{T1} & k_{T2} & \cdots & k_{Tn} \\
% (y_1 - y_{\text{CoG}})k_{T1} & (y_2 - y_{\text{CoG}})k_{T2} & \cdots & (y_n - y_{\text{CoG}})k_{Tn} \\
% -(x_1 - x_{\text{CoG}})k_{T1} & -(x_2 - x_{\text{CoG}})k_{T2} & \cdots & -(x_n - x_{\text{CoG}})k_{Tn} \\
% k_{R1} d_1 & k_{R2} d_2 & \cdots & k_{Rn} d_n \\
% \end{bmatrix}
% \end{equation}
\begin{equation}
\mathbf{A} =
\begin{bmatrix}
k_{T1} & \cdots & k_{Tn} \\
(y_1 - y_{\text{CoG}})k_{T1} & \cdots & (y_n - y_{\text{CoG}})k_{Tn} \\
-(x_1 - x_{\text{CoG}})k_{T1} & \cdots & -(x_n - x_{\text{CoG}})k_{Tn} \\
k_{R1} d_1 & \cdots & k_{Rn} d_n \\
\end{bmatrix}
\end{equation}
}

% \begin{equation}
%     \[
% \mathbf{A} =
% \begin{bmatrix}
% k_{T1} & k_{T2} & \cdots & k_{Tn} \\
% (y_1 - y_{\text{CoG}})k_{T1} & (y_2 - y_{\text{CoG}})k_{T2} & \cdots & (y_n - y_{\text{CoG}})k_{Tn} \\
% -(x_1 - x_{\text{CoG}})k_{T1} & -(x_2 - x_{\text{CoG}})k_{T2} & \cdots & -(x_n - x_{\text{CoG}})k_{Tn} \\
% k_{R1} d_1 & k_{R2} d_2 & \cdots & k_{Rn} d_n \\
% \end{bmatrix}
% \]
% \end{equation} }
\yantian{where \(k_{Ti}\) is the thrust coefficient of the $i$-th rotor, $x_i$ and $y_i$ are the coordinates of the $i$-th rotor relative to the body frame, $k_{Ri}$ is the rolling moment coefficient, $d_i$ represents the rotor's rotational direction, and $x_{\text{CoG}}$, $y_{\text{CoG}}$ are the coordinates of the center of gravity and will be updated per step\footnote{While in this work $x_{\text{CoG}}$ and $y_{\text{CoG}}$ are obtained from our simulator, we are aware of methods that can estimate $x_{\text{CoG}}$ and $y_{\text{CoG}}$ in real world, such as \cite{lee2015control,lee2016estimation}. That said, developing on-line parameter estimators for aerial manipulation is still an open problem, and such features can be added in the future version of this work.}.

To obtain the required rotor angular velocities $\omega$, the inverse of the allocation matrix $\mathbf{A}^{-1}$ is calculated and applied to the vector of desired force and torques $[\mathbf{F}, \tau_x, \tau_y, \tau_z ]^T$:

\begin{equation}
    \omega^2 = \mathbf{A}^{-1} \begin{bmatrix}
           \mathbf{F} \\
           \tau_x \\
           \tau_y \\
           \tau_z
         \end{bmatrix}
\end{equation} The squared angular velocities $\omega^2$ are then processed to ensure they are non-negative, followed by normalization if any value exceeds the maximum permissible squared velocity. Finally, taking the square root of these values gives the rotor angular velocities in radians per second.

This dynamic adjustment ensures that our simulator accurately reflects the quadcopter's behavior as its CoG shifts due to varying payloads or changes in configuration, maintaining precise control and stability throughout its operation.

% Aerial manipulator using on-line parameter estimator for an unknown payload is still an open research. While previous works have proposed ways to help estimate control parameters such as center of gravity by merely relying on acceleration data, we believe our simulator would serve as a useful tool to foster such research. 

\subsection{Control Development} \label{sec:control_dev}

The block diagram showing the AAM control system is presented in Fig.~\ref{fig:block_diagram}. The desired velocity and the desired joint angles for the quadrotor and manipulator, respectively, are calculated using the PID (Proportional-Integral-Derivative) and PD (Proportional-Derivative) controllers.

A PID controller was designed to regulate the velocity state of the drone, drawing inspiration from the work presented in \cite{5980409}. The performance of the PID controller indicated relatively good attitude stabilization. Equ. \ref{eq:pid_eq}. was employed to compute the control force using the PID controller for the quadrotor. This force was then allocated to individual rotors to determine their respective angular velocities, as mentioned in Sec.~\ref{sssec:AAMDM} and similar to this work \cite{5980409}.

% \begin{figure}[H]
%     \centering
%     \includegraphics[width=1\linewidth]{figures/block diagram for controller.png}
%     \caption{Block Diagram of Controller for AAM}
%     \label{fig:block_diagram}
% \end{figure}

% GIVE AN OVERVIEW (Put a figure)
\begin{figure*}[t]
    \centering
    \includegraphics[width=1\linewidth]{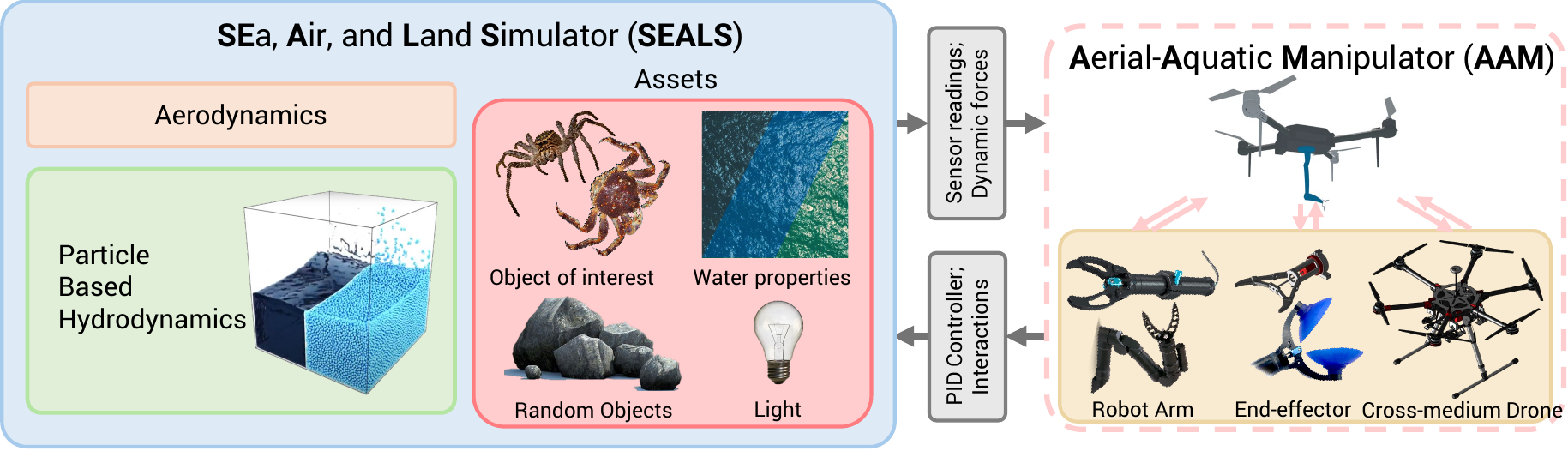}
    \vspace*{-6mm}\caption{Overview of our SEa, Air, and Lands Simulator (SEALS).}
    \label{fig:overview}\vspace*{-4mm}
\end{figure*}

\begin{equation} \label{eq:pid_eq}
    F = K_pE_p + K_dE_d + K_iE_i + [0, 0, m_1g] + m_1a_{ref_{1}} 
\end{equation}
where 
\setlist{nolistsep}
    \begin{itemize}[noitemsep]
        \item $F$ is the force applied to Quadrotor. 
        \item $K_pE_p$ is the Proportional term of PID Controller. $K_p$ is the Proportional Gain and $E_p = v - v_{ref}$, is an error between drone velocity$(v)$ and desired or reference velocity$(v_{ref})$.
        \item $K_dE_d$ is the Derivative term of PID Controller. $K_d$ is the Derivative Gain and $E_d = (v - v_{prev})/d_t - a_{ref_1}$, is an error between drone acceleration $((v - v_{prev})/d_t)$ and desired or reference acceleration$(a_{ref_1})$, and $d_t$ is the time step.
        \item $K_iE_i$ is the Integral term of PID Controller. $K_i$ is the Integral Gain and $E_i$ is the cumulative summation of $E_p$ at each time step.
        \item $[0, 0, m_1g]$ is the gravitational force acting on the quadrotor.
        \item $m_1a_{ref_1}$ is the force acting on the quadrotor, where $m_1$ is the mass of the quadrotor, and $a_{ref_1}$ is the reference acceleration, which is the desired acceleration of the quadrotor. 
        \item Dimensions for $v, v_{ref}, $ and $ a_{ref_1}$ represent the $x,y,z$ directions in 3D space, denoted as $\mathbb{R}^3$. 
    \end{itemize}

The manipulator joints are controlled using the PD controller described by Equ. \ref{eq:pd_eq}. The function `set\_dof\_target\_pos()', in Isaac Sim, is employed to define the target joint angle positions for the manipulator, which calculates the desired velocity and angle for the manipulator.

\begin{equation} \label{eq:pd_eq}
    F^m = K^m_pE^m_p + K^m_dE^m_d
\end{equation}
where 
    \begin{itemize}
        \item $F^m$ is the force applied to a manipulator joint.
        \item $K^{m}_{p}E^m_p$ is the proportional term of the PD controller, where $K^m_p$ proportional gain of the joint and $E^m_p = x_{ref} - x $ is the error between the desired (reference) angular position $x_{ref}$ and the current angular position $x$ of the joint.
        \item $K^m_dE^m_d$ is the derivative term of the PD controller, where $Km_d$ is the Derivative Gain and $E^m_d = (x - x_{prev})/d_t$, with $x_{prev}$ being the previous angular position and $d_t$ the time step.
    \end{itemize}
    
}

% \begin{figure}
%     \centering
%     \includegraphics[width=\textwidth]{figures/pid_controller_drone.png}
%     \caption{PID Controller}
%     \label{fig:pid_controller}
% \end{figure}

\section{\textcolor{Aquamarine}{\textbf{SE}}a, \textcolor{Aquamarine}{\textbf{A}}ir, and \textcolor{Aquamarine}{\textbf{L}}and \textcolor{Aquamarine}{\textbf{S}}imulator (\textcolor{Aquamarine}{\textbf{SEALS}})}\label{sec:SEALS}

\subsection{Aerodynamics Development}
Similar to the approach adopted by Jacinto et al. in \cite{jacinto2023pegasus}, a simplified linear drag force model is employed to represent the aerodynamic effects that act  on the drone. The influence of this linear drag force on our AAM can be expressed using the following equation (Equation~\ref{eq:linear_drag}):
\begin{equation} \label{eq:linear_drag}
  F_d = c \mathbf{v}
\end{equation}
where:
\begin{itemize}
    \item $F_d$ denotes the drag force with units of $N$ (dimension $\mathbb{R}^3$).
    \item $\mathbf{v} = [\dot{x} \quad \dot{y} \quad \dot{z}]^T$ represents the linear velocity of the body frame ($F_B$) with respect to the world frame ($F_I$).
    \item $c$ is a constant vector with units of $N/(m/s)$ (dimension $\mathbb{R}^3$), representing the drag coefficient dependent on the velocity acting on the body along each axis. Each element of $c$ lies within the range $[0, 1)$.
\end{itemize}

\subsection{Underwater Dynamics Development} \label{sec:hydrodynamics}
% \vspace*{-3mm}

One of the most challenging aspects of building a high-fidelity simulator that features an underwater domain is accurately modeling hydrodynamics and hydrostatics. Water particles behave in complex, often unpredictable, movement and collision. As such, modeling the forces acting in resistance to underwater motion of a rigid body cannot be accurately calculated by rigid body hydrodynamic equations as used in simulators such as MARUS \cite{lonvcar2022marus} and UNav-Sim \cite{amer2023unav} across various underwater environments. To increase the fidelity of the underwater simulation, smoothed particle hydrodynamics (SPH) \cite{monaghan1992smoothed} has been used to simulate the behaviors of individual fluid particles and how they interact with each other and the environment. This method is particularly viable for representing complex fluid interactions such as oceanography, currents, waves, and boundary conditions concerning hydrodynamics \cite{fraga2019smoothed,xi2019survey,zago2017simulating}. In addition, there are a variety of applications in which SPH excels, including computational biology \cite{toma2021fluid}, simulation of underwater landslides \cite{zago2017simulating}, and modeling of ice formations in a sea \cite{marquis2022smoothed}. 

The variety of useful and high-fidelity applications of SPH makes it an attractive choice to model hydrodynamics for SEALS, but there are some core stability and computational issues, as noted by Macklin and Müller \cite{muller2007position}. To address this, Macklin and Müller introduced a method titled position-based dynamics (PBD). This technique incorporates SPH, but introduces a constant density constraint that enforces particle incompressibility, allowing for longer timesteps in calculation and better performance when scaled \cite{muller2007position,andersson2023comparison}. It is for these reasons that we chose Isaac Sim's PhysX engine to simulate high-fidelity hydrodynamics using PBD \cite{macklin2014unified}. This system gives SEAL a strong and cutting-edge balance of realistic dynamics, breadth of application, and computational efficiency, all of which will only increase as hardware improves. We have included an overview of the hydrodynamics in Appendix.~\ref{Ap:PBD}.
% oceanography \cite{xi2019survey,fraga2019smoothed,zago2017simulating}

% Simulating complex fluids with smoothed particle hydrodynamics \cite{zago2017simulating}

% fluid–structure interaction Biological Systems \cite{toma2021fluid}

% submarine \cite{wang2016numerical}

% Sea-Ice Model \cite{marquis2022smoothed}

% Why uses of particle-dynamics in Issac Phyx is promising? Make sure this is convincing. Comparison can be made with other potential hydrodynamics methods, such as UNav-sim. This is fundamentally important to argue for our contribution in building the simulator on NVIDIA Issac Sim.

% compare particle-based and position-based hydrodynamics  \cite{andersson2023comparison}

\yantian{
% \vspace*{6mm}

\subsection{Simulation Realism} \label{sec:realism}
% \vspace*{-10mm}

The underwater part of our SEALS has the unique feature of enhancing realism as follows.

\begin{figure*}[!tbh] % Use 'b' for bottom placement
    \centering
    \includegraphics[width=1\linewidth]{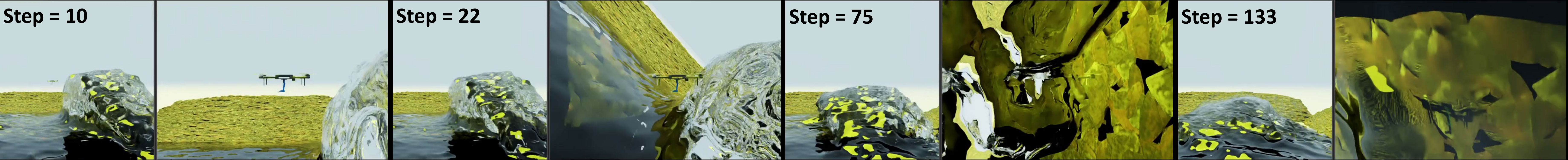}
    \vspace*{-6mm}\caption{Video frames of our Aerial-Aquatic Manipulator entering the water while enduring an ocean wave, with water damping and light refraction effects.}
    \label{fig:wave_vis}\vspace*{-5mm}
\end{figure*}

\subsubsection{Realistic Air-Water Transition}

Our SEALS system uses particle-based hydrodynamics to achieve highly realistic air-water transitions, capturing both dynamic interactions and detailed rendering. As discussed in Appendix \ref{Ap:PBD}, this approach simulates cohesion and surface tension, allowing realistic interactions between fluid and solid surfaces. Grounded in solid theoretical principles, we observe that particle-based hydrodynamics in SEALS effectively simulate water splashes when the AAM impacts the water and damping effects as it transitions from air to water, while the damping effect is demonstrated by deactivating the AAM's thrusters and allowing it to descend into the water under gravity. This effect causes a sudden change in acceleration as the AAM enters the water. Due to disturbances in the surface of natural water, such as wind-induced waves, the AAM loses the balance it maintains in the air once it submerges. %For a detailed demonstration, please refer to the accompanying video: \url{https://shorturl.at/XG9Cr}

\subsubsection{Realistic Wave-Drone Interaction}

The causes of ocean waves are diverse and winds can also vary significantly. While wind is the primary driver, generating waves by transferring energy to the water's surface, other factors also play a role. Seismic activity, such as underwater earthquakes, volcanic eruptions, or landslides, can produce tsunamis that may reach heights exceeding 100 feet (30 meters) in extreme cases. Additionally, the gravitational pull of the moon and sun creates tidal waves, which are typically more gradual and predictable compared to wind-driven waves and tsunamis. Therefore, simulating water waves in a controllable way is an important feature of our SEALS to enhance realism. 

In the Fig.~\ref{fig:wave_vis}% and this demo video: \url{https://shorturl.at/aT8yf}
, we showcase an AAM in free fall that is unexpectedly struck by an ocean wave. The sudden shifts in acceleration caused by the wave impact are quantitatively illustrated in Fig.~\ref{fig:Wave_Acceleration}.

% However, particle-based hydrodynamics simulation is very expensive and it is challenging to simulate water area that's large enough so that the accumulative effects of wind

\subsubsection{Realistic Underwater Observation}

In underwater environments, light attenuation significantly impacts visibility due to absorption and scattering by water molecules and suspended particles. As depth increases, light intensity diminishes, leading to reduced visibility, as shown in Fig.\ref{fig:Light_Intensity_Control}. 
\vspace*{-3mm}
\begin{figure}[H]
    \centering
    \includegraphics[width=1\linewidth]{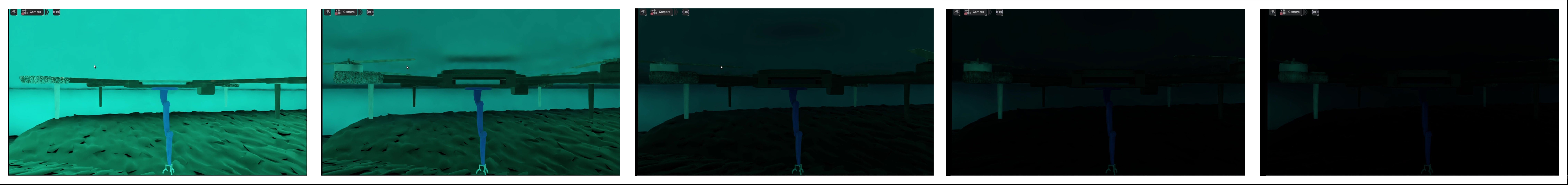}
    \vspace*{-6mm}\caption{Video frames showing light attenuation when the robot dives deeper. }
    \label{fig:Light_Intensity_Control}\vspace*{-1mm}
\end{figure}

Additionally, different wavelengths are absorbed at varying rates; longer wavelengths like red are absorbed more quickly, while shorter wavelengths like blue penetrate deeper, causing distant objects to appear bluer and more blurred. This depth-dependent color shift and blurriness, illustrated in Fig.~\ref{fig:Wavelen_Absorp_Effect} (middle), affect how objects are perceived underwater. Furthermore, light scattering, often observed as caustics, occurs when light rays bend and disperse through varying water densities, creating intricate patterns of light and shadow on submerged surfaces, as shown in Fig.~\ref{fig:Wavelen_Absorp_Effect} (right).

\vspace*{-3mm}
\begin{figure}[H]
    \centering
    \includegraphics[width=1\linewidth]{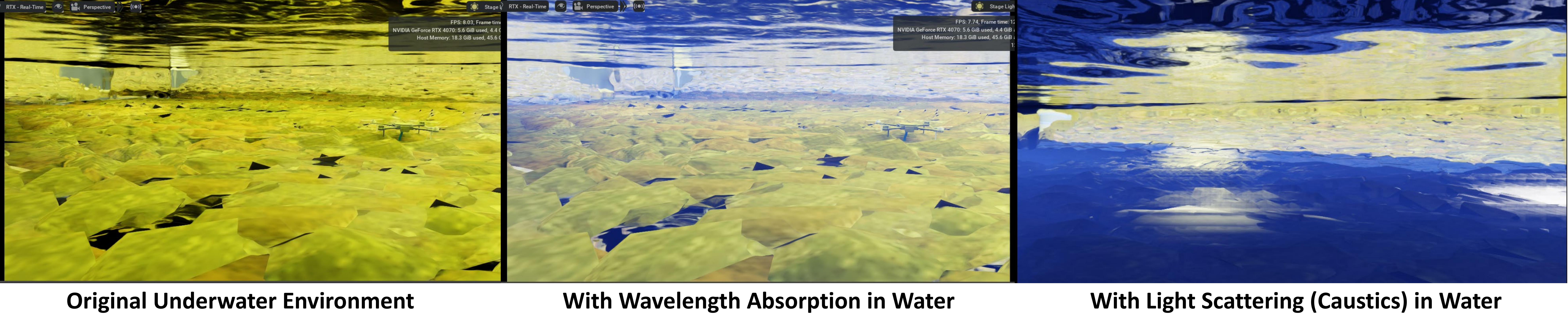}
    \vspace*{-6mm}\caption{Visual representations of an underwater environment illustrating the effects of wavelength absorption, and light scattering (caustics).}\vspace*{-3mm}
    \label{fig:Wavelen_Absorp_Effect}
\end{figure}

\subsubsection{Realistic Aquatic Animals}\label{sec:realism:animals}

While most photorealistic simulators focus primarily on sensory realism, our SEALS system is the first to also emphasize the realistic simulation of environmental animal behavior. We have meticulously developed detailed meshes and kinematic models of aquatic animals, such as crabs and sea spiders, and equipped them with controllers that facilitate robot learning for developing control policies, as shown in Fig.~\ref{fig:Crab}. 
\vspace*{-3mm}
\begin{figure}[H]
    \centering
    \includegraphics[width=1\linewidth]{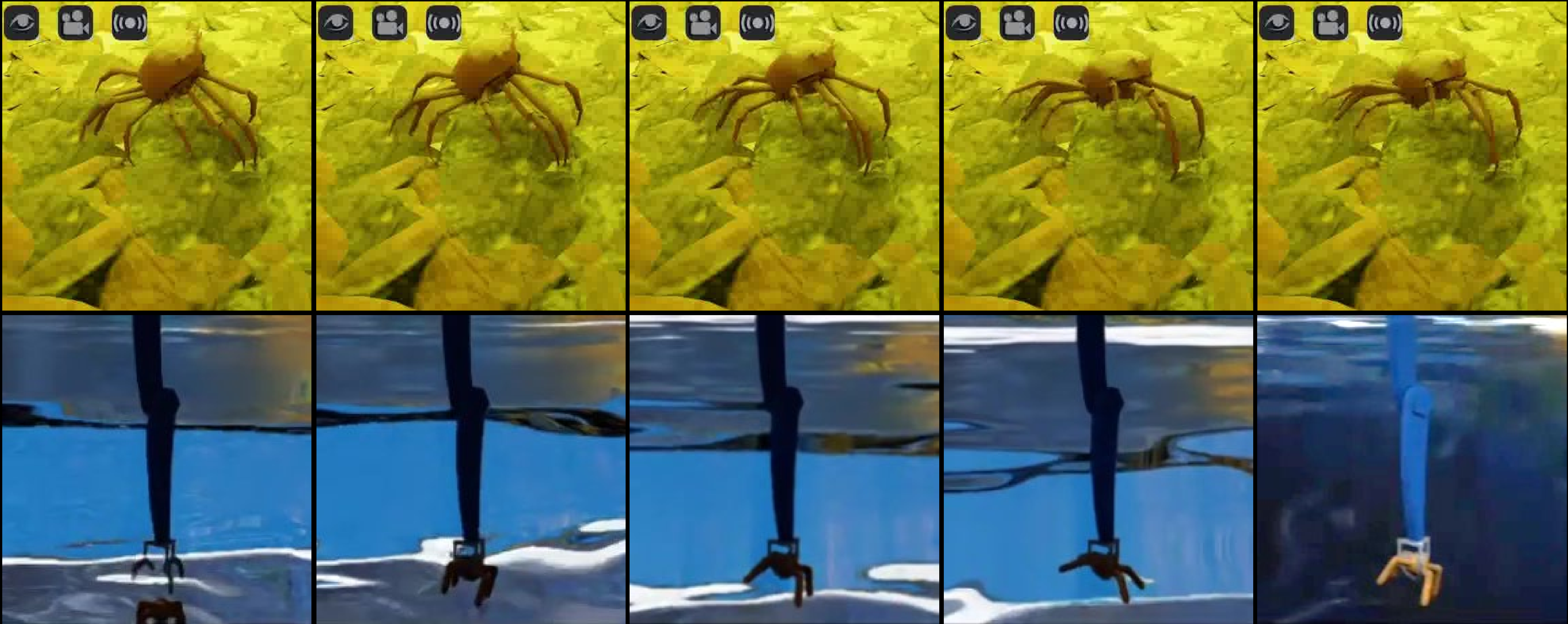}
    \vspace*{-6mm}\caption{\textbf{Top:} Simulated crab walking slowly on the sea floor; \textbf{Bottom:} Simulated sea spider captured by our Aerial-Aquatic Manipulator.}
    \label{fig:Crab} \vspace*{-3mm}
\end{figure}

% \begin{figure}[H]
%     \centering
%     \includegraphics[width=1\linewidth]{figures/SeaSpider.pdf}
%     \vspace*{-6mm}\caption{Simulated sea spider captured by our Aerial-Aquatic Manipulator.}
%     \label{fig:SeaSpider}\vspace*{-4mm}
% \end{figure}

This behavioral realism is crucial for practical applications in in-land aquaculture assistance, marine biology sampling, and fish farming, as shown in Fig.\ref{fig:intro}. By creating digital replicas of real-world animals, SEALS allows for more realistic and effective training of the Aerial-Aquatic Manipulator (AAM). However, achieving realistic aquatic animal simulations is challenging, requiring the construction of detailed meshes, accurate segmentation into parts, precise joint definitions to enable realistic movements, and the integration of joint controllers with reinforcement learning to develop control policies. We provide guidelines in Appendix \ref{Ap:Aquatic_Animals} on how we accomplished this.
}

\subsection{Sensors and Perceptual Modalities for Robots}
In this initial version of AAM-SEALS, We implemented the following sensors:

% \begin{figure}[H]
%     \centering
%     \includegraphics[width=1\linewidth]{figures/Aerial_Aquatic_Manipulator_Figs_cameras_22.pdf}
%     \caption{Camera Positions (Red, Green, and Blue Arrows are X, Y, and Z Axes, Respectively) and Views of the Cameras}
%     \label{fig:camera_setting}
% \end{figure}
\vspace*{-3mm}
\begin{figure}[H]
    \centering
    \includegraphics[width=1\linewidth]{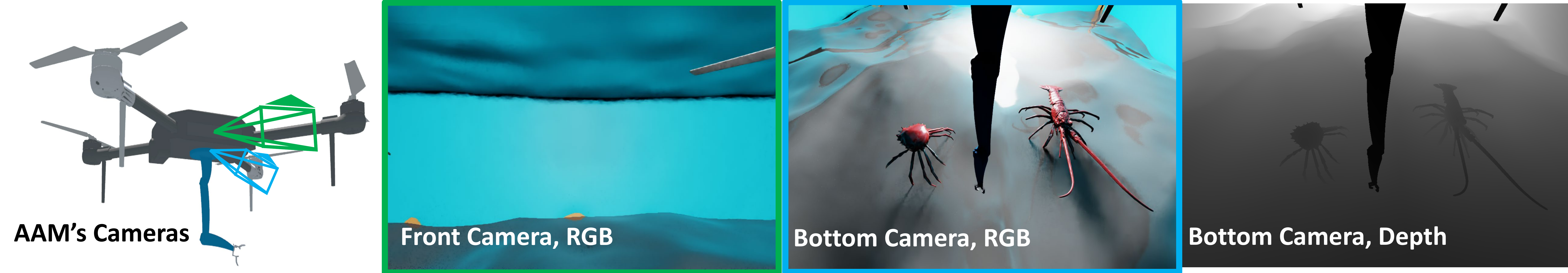}
    \vspace*{-6mm}\caption{Camera positions (red, green, and blue arrows are X, Y, and Z axes, respectively) and views of the cameras}
    \label{fig:camera_setting}\vspace*{-3mm}
\end{figure}

% This section details the robot's simulated perceptual modalities used to train the reinforcement learning (RL) agent within Isaac Sim. The agent, modeled as a drone, leverages two primary sensors, both made available through Isaac Sim's rendering and physics capabilities: 

\mund{Camera Sensors} The drone is equipped with two RGB-Depth camera sensors, as depicted in Fig.~\ref{fig:camera_setting}. The first is a front-facing camera mounted on the side of the drone, providing a forward view. The second is a downward-facing camera located on the belly near the edge of the drone, designed to capture a view with maximum overlap of the manipulator workspace. %For RL training, we focused on a depth image from the down-facing camera and an RGB image from the front-facing one.

\mund{Contact Sensor} This sensor detects physical contact between the gripper attached to drone manipulator and other rigid bodies in the environment. When a force exceeding a predefined threshold is applied to the body where the sensor is attached, the sensor transmits a signal indicating contact. The Contact Sensor extension utilizes the PhysX Contact Report API to generate a reading comparable to real-world contact cells or pressure sensors. For this experiment, contact sensors were positioned at each gripper fingertip.

\subsection{Control Interfaces to Robots}\label{sec:SEALS:Control}

The control interface between Reinforcement Learning and the Isaac Simulator involves a system in which the application sends commands for quadrotor velocity, joint angles, and gripper action to the Actions module. These actions are managed by the Controller module, which includes a PID Controller for quadrotor velocity, a PD Controller for manipulator joint angles as mentioned in Sec.~\ref{sec:control_dev}, and a gripper command module. The PID Controller calculates the quadrotor's velocity, simulates the drone dynamics, converts force on the drone into rotor forces using the Allocation Matrix, and calculates rotor torques. The PD Controller manages the manipulator dynamics by determining joint angles, while the gripper command controls the open/close actions. These outputs are sent to the Drone Dynamics and Manipulator Dynamics models, which process the physical behavior of the drone and manipulator. Finally, the results are sent to the NVIDIA Isaac Sim, enabling real-time simulation and control of the drone and manipulator.

\subsection{Aquatic Animal Search and Capture Challenge for AAMs}
Building upon the capabilities of our SEALS simulator, which features fully controllable models of a crab and a sea spider (as detailed in Sec.~\ref{sec:realism:animals}), we present a novel challenge: the search and capture of aquatic animals. This task demands intricate tacit strategies and the flexibility to adapt learned skills to unforeseen scenarios.

To facilitate the transfer of these implicit strategies to robots, we employ teleoperation of our AAM using a joystick to gather demonstrations for search-and-capture tasks. We advocate for the use of reinforcement learning from demonstrations (RLfD), a method that merges the efficiency of acquiring expert knowledge through imitation learning with the exploratory strengths of reinforcement learning, as shown in prior works \cite{nair2018overcoming, zha2024learning}. Typically, RLfD requires as few as 10 demonstrations to achieve effective learning. Our demonstrations are available at this link: \url{https://tinyurl.com/5b8sm9fc}.%and corresponding AAM trajectories at: \url{https://shorturl.at/pzski}.

\subsection{Reinforcement Learning from Demonstrations (RLfD)}
Building on the control interfaces explained in Sec.~\ref{sec:SEALS:Control}, we implement our Reinforcement Learning from Demonstrations (RLfD) framework. We model the environment as a Markov Decision Process (MDP) defined by the tuple $(S, A, P, R, \gamma)$, where $S$ denotes the state space, $A$ the action space, $P(s'|s,a)$ the state transition probability, $R(s,a)$ the reward function, and $\gamma$ the discount factor. Our RLfD implementation builds upon Soft Actor-Critic from Demonstrations (SACfD) implemented in \cite{zha2024learning}, which integrates expert demonstrations into Soft Actor-Critic (SAC) \cite{haarnoja2018soft} to accelerate policy learning while maintaining exploration. SAC is a state-of-the-art RL algorithm for continuous control tasks, which optimizes policies by maximizing both the expected cumulative reward and an entropy term to encourage exploration, our approach integrates expert demonstrations to expedite learning. The objective function in SAC is given by:

\begin{equation}
J(\pi) = \mathbb{E}_{\pi} \left[ \sum_{t=0}^{\infty} \gamma^t \left( R(s_t, a_t) + \alpha \mathcal{H}(\pi(\cdot|s_t)) \right) \right]
    % J(\pi)=\sum_{t=0}^{\infty} \mathbb{E}_{s_t,a_t \sim \rho_\pi} [r_{s_t,a_t}+\alpha \mathbb{H}(\pi(\cdot|s_t))]
\end{equation} where $\alpha$ is a temperature parameter that balances the reward and entropy terms, and $\mathcal{H}(\pi(\cdot|s_t))$ represents the entropy of the policy at state $s_t$. To effectively incorporate demonstrations, we employ several strategies:

\begin{itemize}
    \item Pretraining: SACfD initializes the agent's policy and value functions by pretraining on the demonstration data, which provides a strong starting point before interacting with the environment.
    \item Prioritized Experience Replay: SACfD utilizes a prioritized replay buffer where demonstration experiences are assigned higher priorities, ensuring that the agent samples these informative transitions more frequently during training.
    \item N-Step Returns: To better learn longer-term dependencies from demonstrations and provide more informative updates, SACfD incorporates n-step returns in the learning process, which has the following form:
\begin{equation*}
G_t = \sum_{k=0}^{N-1} \gamma^k R_{t+k} + \gamma^N V(s_{t+N})
\end{equation*} where $V(s_{t+N})$ is the estimated value of the state after $N$ steps. $R_{t+k}$ is the reward at step $t+k$.
\end{itemize}

% \vspace*{-3mm}
\begin{figure*}[t]
    \centering
    \includegraphics[width=1\linewidth]{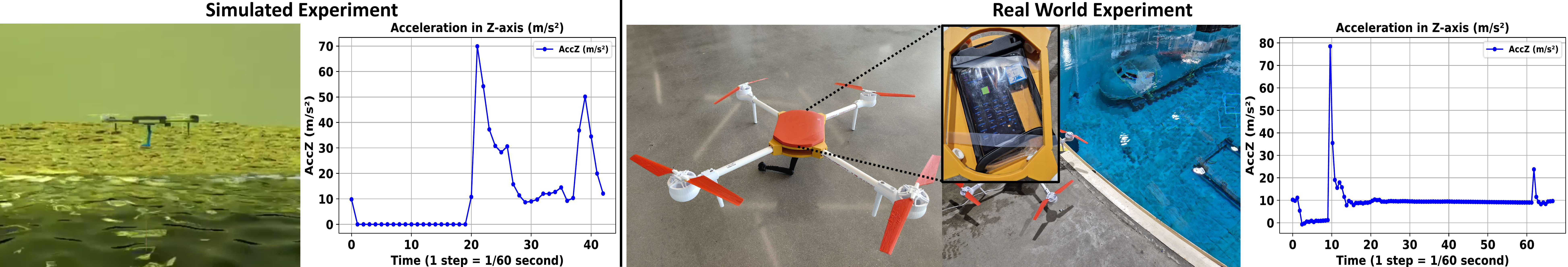}
    \vspace*{-7mm}\caption{\textbf{Left:} A simulated AAM and its acceleration along the z-axis over time as it falls from air into water. \textbf{Middle:} The 3D-printed AAM and the water tank used in real-world experiments. \textbf{Right:} Acceleration along the z-axis over time for the real-world AAM falling into water.}
    \label{fig:air_water_acceleration}\vspace*{-1mm}
\end{figure*}

\begin{figure*}[t]
  \centering
  \includegraphics[width=1\textwidth]{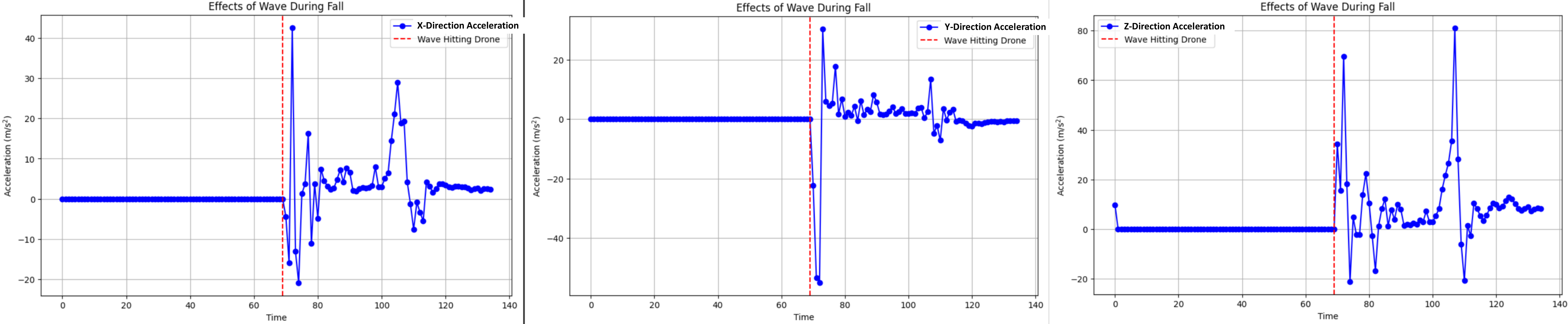}
  \vspace*{-8mm}\caption{Plot of acceleration along x, y, and z directions over time while AAM falls freely from the air and gets hit by a wave.}
  \label{fig:Wave_Acceleration}\vspace*{-5mm}
\end{figure*}

\section{Evaluation}
\label{section:Evaluation}

Our AAM-SEALS system includes both the robot and the simulator, guiding our evaluation to \yantian{focus on three major aspects: simulation, robot, and application. 

\noindent\textbf{Simulation:} we focus on quantitatively assess the fidelity of the particle-based underwater dynamics by: 1) obtaining acceleration-over-time curves for various objects, including the AAM, freely dropping into water in the SEALS simulator; 2) comparing these curves with real-world counterparts, where objects equipped with Inertial Measurement Unit (IMU) sensors were dropped into a water tank.%; and 3) implementing position-based water dynamics from the recently published photorealistic underwater simulator, UNav-Sim \cite{amer2023unav}, and conducting a comparative analysis of position tracking between particle-based and position-based dynamics. 

\noindent\textbf{Robot:} We evaluate the effectiveness of our AAM's control by analyzing its position-tracking error, a fundamental metric for assessing the accuracy and performance of control systems. This analysis highlights the system's ability to follow predefined trajectories with precision, ensuring reliable operation in diverse scenarios. %we aim at evaluating the effectiveness of control of our AAM by: 1) showcasing the position-tracking error, a major metrics for evaluating the control of unmanned aerial vehicles \cite{}; 2) 

\noindent\textbf{Application:} We conducted visual reinforcement learning experiments within AAM-SEALS using state-of-the-art techniques, such as Soft Actor-Critic (SAC), to demonstrate its effectiveness for robot learning tasks. %We conducted visual reinforcement learning within AAM-SEALS using cutting-edge techniques such as reinforcement learning from demonstrations, demonstrating its suitability for robot learning, for multiple tasks, navigation and crab searching and capturing. 

% \vspace*{-4mm}
\subsection{Evaluating the Realism of Particle-based Hydrodynamics}\label{sec:PBD_Rigid}
% \vspace*{-1mm}

Given the complexities of achieving high-fidelity hydrodynamics, it is crucial to evaluate the realism of particle-based hydrodynamics implemented within our SEALS simulator. Due to the differences between the densities of air and water mediums, the realistic hydrodynamics would cause damping effects to objects operating in the water. %Following this insight, we evaluate hydrodynamics by comparing the acceleration changes of objects freely dropped into water in the real world and their counterparts in our SEALS simulator. All of the objects are equipped with an IMU sensor. We considered a broad range of objects including various containers and a 3D-printed AAM\footnote{We used 3D-printed AAM due to the extreme difficulty of building a fully functional physical AAM (see the Limitation in Sec. \ref{sec:conclusion} for details.}. We showcase those objects and our water tank facility in Fig.~\ref{}.
Following this insight, we evaluate hydrodynamics by comparing the acceleration changes of objects freely dropped into water in the real world and in our SEALS simulator. All of the objects are equipped with an IMU sensor. 

% The Fig.~\ref{} exhibits the acceleration-over-time curves of those objects and their counterparts in our SEALS simulator. From the figure we observe that our hydrodynamcis is close to real world hydrodynamics, both featuring a sudden shift in acceleration at the moment the AAM enters the water, caused by water damping effects.
\xiaomin{
In SEALS, we disabled the AAM's thrusters and allowed it to fall freely from the air into the water. For the real-world experiment, we equipped a 3D-printed AAM with an IMU sensor (WitMotion WT901BLECL, 50 Hz output) and dropped it into a water tank. We recorded the changes in the AAM's acceleration over time for both simulated and real world experiments, as shown in Fig.~\ref{fig:air_water_acceleration}. The results exhibit similar patterns in both cases: the acceleration initially starts at gravitational acceleration, then rapidly decreases to near zero due to air resistance. Upon water entry, a sharp fluctuation occurs due to water damping effects, followed by another fluctuation when the AAM hits the ground.
}

}

\begin{figure*}[tbh]
  \centering
  \includegraphics[width=1\textwidth]{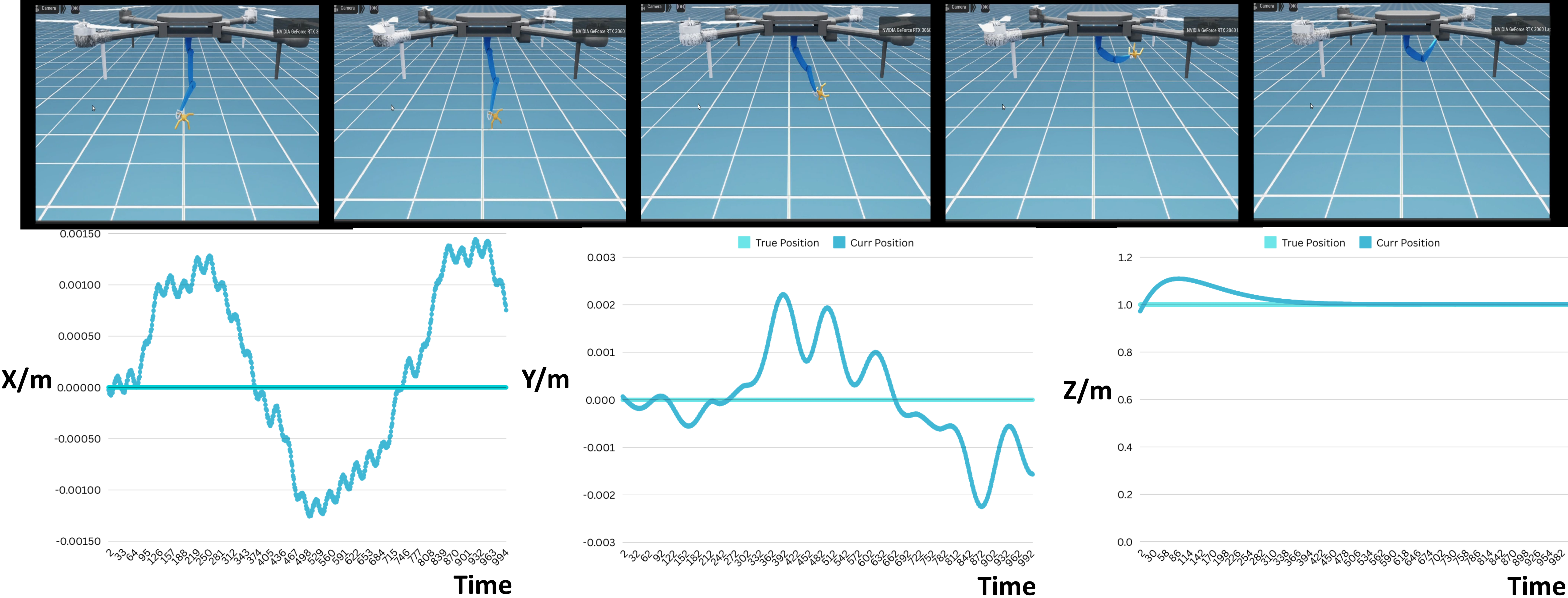}
  \vspace*{-6mm}\caption{\textbf{Top:} Sample video frames showcasing the AAM hovering with a moving manipulator; \textbf{Bottom:} Position-tracking curves over time for the X, Y, and Z directions.}\vspace*{-1.8mm}
  \label{fig:hovering}
\end{figure*}

\begin{figure*}[tbh]
  \centering
  \includegraphics[width=1\textwidth]{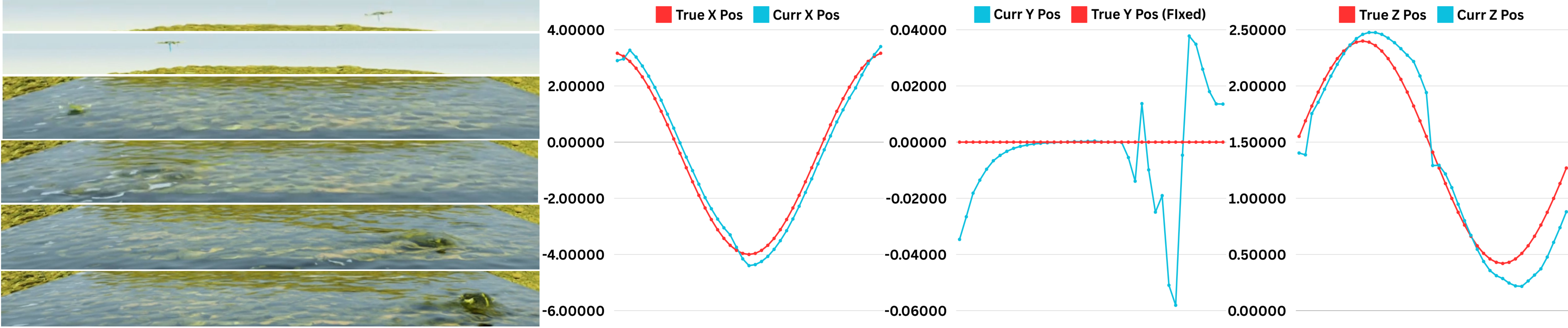}
  \vspace*{-7mm}\caption{Oval trajectory following results.}\vspace*{-3mm}
  \label{fig:oval}
\end{figure*}

% \vspace*{-6mm}
Finally, Fig.~\ref{fig:Wave_Acceleration} illustrates the significant shifts in accelerations along the x, y, and z axes when the AAM was impacted by an ocean wave, further demonstrating the dynamic realism of our hydrodynamics simulation.

\yantian{
\subsection{Evaluating the Control of AAM in SEALS Simulator}
We evaluate the AAM's control performance by analyzing position-tracking error across multiple tasks: \textbf{Task 1} involves hovering control with arm movements to evaluate the robustness of the system to changes in the center of gravity, while \textbf{Task 2} focuses on following an oval trajectory that transitions between air and water, testing the control architecture's cross-medium effectiveness.

\xiaomin{
The position-tracking results of hovering task, shown in Fig.~\ref{fig:hovering}, demonstrate precise control while being robust to the changes of Center of Gravity: the X-axis remains within $\pm 0.015 m$, the Y-axis within $\pm 0.003m$, and the Z-axis within $\pm 0.2m$. Likewise, the oval (cross-medium) trajectory following results, as shown in Fig.~\ref{fig:oval}, also demonstrates the effectiveness of our controller. %The demo video can be visualized at: \url{https://aam-seals.github.io/aam-seals-v1/media/videos/hovering_demo1.mp4}
}
}

\xiaomin{
\subsection{Reinforcement Learning from Demonstration Evaluation}
Fig.~\ref{fig:RLfD_results} demonstrates successful training of RLfD in our SEALS simulator for our AAM robot. 

\begin{figure}[t] \vspace*{-0.1cm}
  \begin{center}
    \includegraphics[width=1\linewidth]{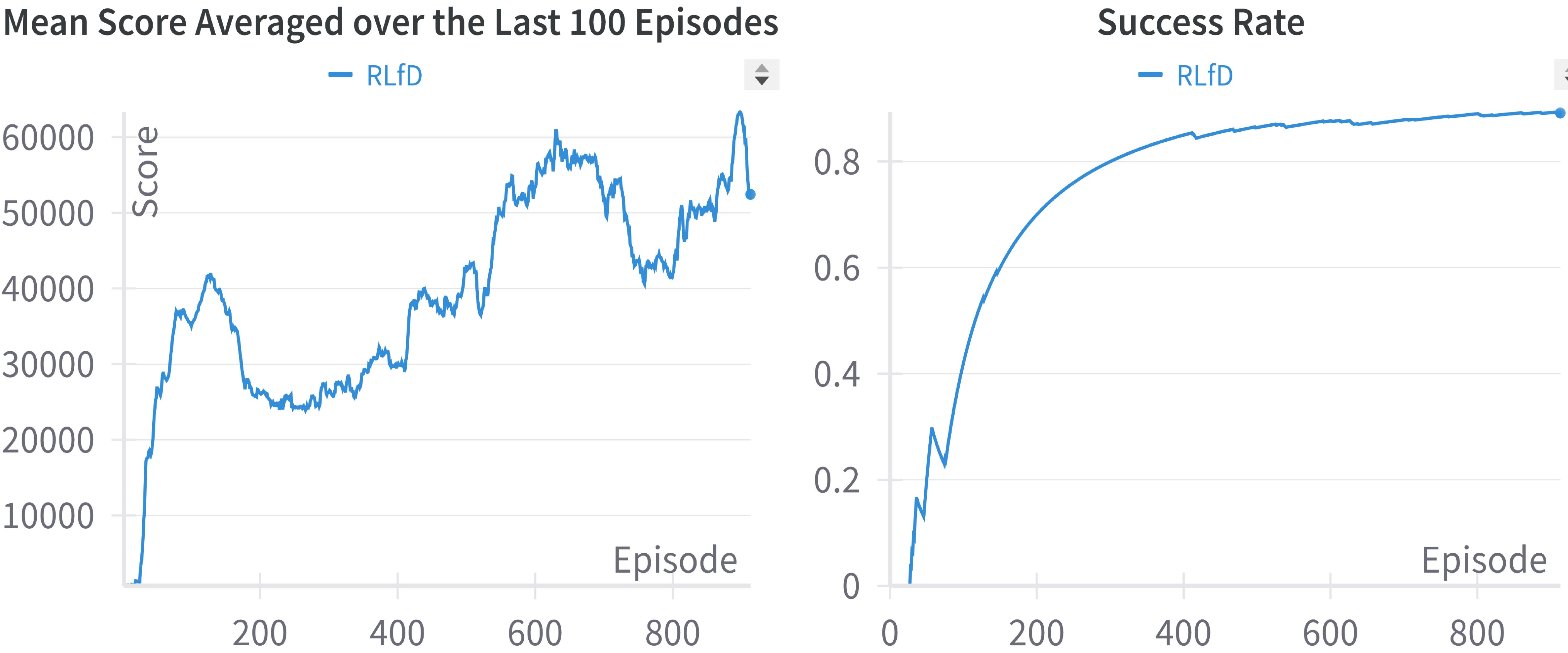}
  \end{center}  
  \vspace*{-5mm}\caption{Reinforcement Learning from Demonstration results}\vspace*{-5mm}
  \label{fig:RLfD_results}
\end{figure}
}

\subsection{Visual Reinforcement Learning Evaluation}
% description of the task: learn to capture crabs/lobsters

\begin{figure}[t] 
  \begin{center}
    \includegraphics[width=0.3715\textwidth]{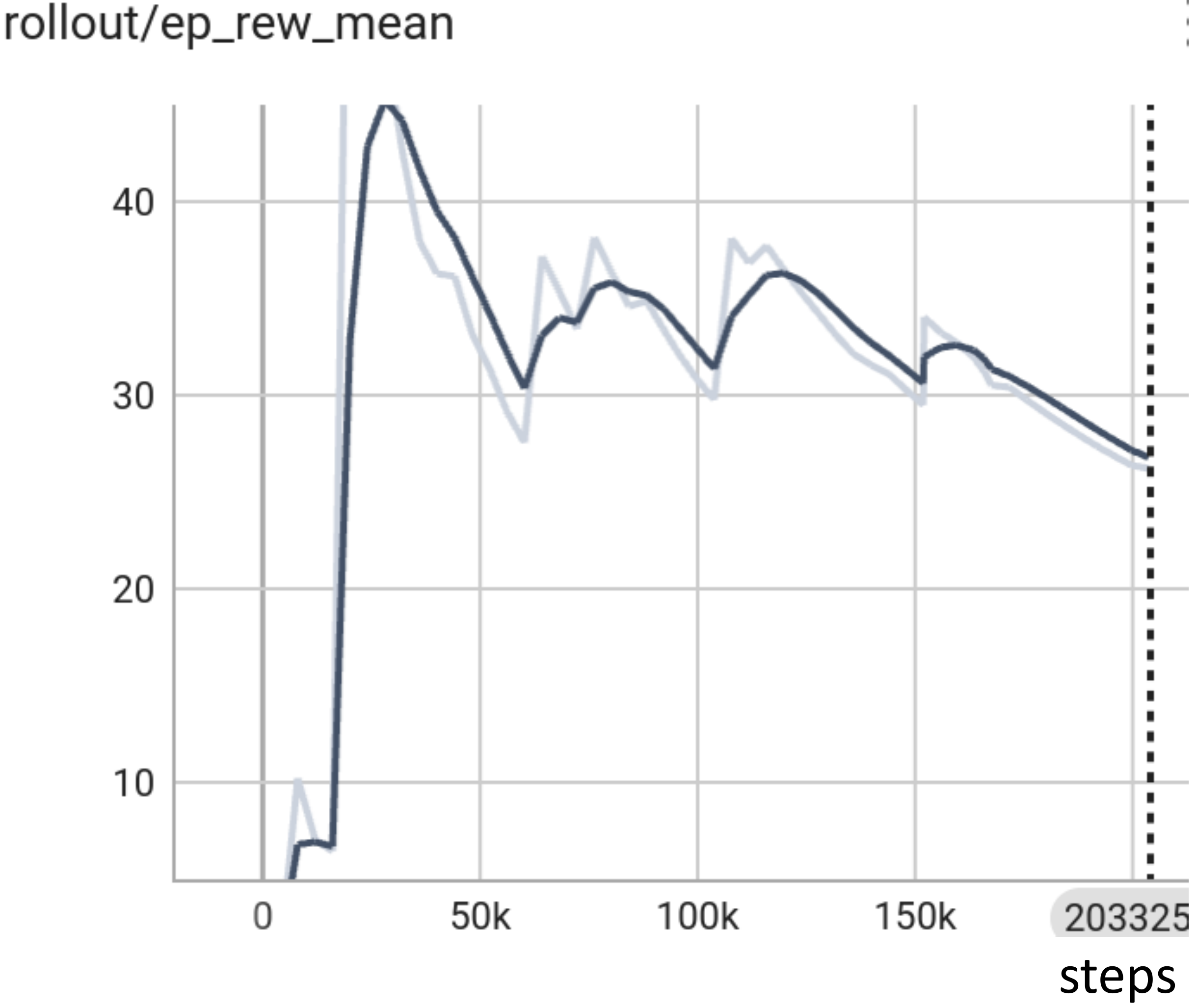}
  \end{center}
  \vspace*{-5mm}\caption{Visual reinforcement learning results}
  \label{fig:vis_RL}\vspace*{-5mm}
\end{figure}

We applied the Soft Actor-Critic algorithm \cite{haarnoja2018soft}, a robust reinforcement learning (RL) method for continuous control, to train our AAM in the SEALS environment to reach objects such as crabs in Fig.~\ref{fig:camera_setting}, using distance-based rewards. The observation space included two 128 x 128 RGB images from the front and bottom cameras on the AAM (resulting in six channels in total) and the global poses of the AAM and the target object. The cumulative average rewards over the past 100 episodes, shown in Fig.~8, indicate that the reinforcement learning converges. The learning results is shown in Fig.~\ref{fig:vis_RL}. For more detailed robot learning results, please refer to Appendix \ref{Ap:RL}.

\vspace*{-0.9cm}
% We applied Soft-Actor Critic algorithm \cite{haarnoja2018soft}, a powerful RL for continuous control, to train our AAM in SEALS to reach to an object, such as a crab \cite{fig:camera_setting}, using distance-based rewards. We considered a hybrid observation space with two 128 x 128 RGB images from the front and bottom cameras on our AAM (6 channels in total), and the global poses AAM and object to be grasped. The episode cumulative rewads averaged over past 100 episodes is shown in Fig.~\ref{fig:vis_RL}, showing that the reinforcement learining converges. For more robot learning results, please refer to Appendix.~\ref{Ap:RL}.

% \begin{wrapfigure}[13]{r}{0.3715\textwidth} \vspace*{-0.1cm}
%   \begin{center}
%     \includegraphics[width=0.3715\textwidth]{figures/RL_1.pdf}
%   \end{center}
%   \label{fig:vis_RL}\vspace*{-5mm}
%   \caption{Visual reinforcement learning results}
% \end{wrapfigure}

\vspace*{7mm}
\section{CONCLUSION}
\label{sec:conclusion}

Our work makes significant contributions to the field of robotics by introducing a new class of robots, Aerial-Aquatic Manipulators (AAMs), and developing the first high-fidelity simulator that integrates sea, air, and land environments. The unique benefits of this work include enabling AAMs to seamlessly transition between different environments and perform complex tasks that require cross-medium manipulation.

\textbf{Limitations:} While our research presents a robust simulation environment, we acknowledge the limitation of not being able to fully verify the Sim2Real transfer. Developing a fully functional physical AAM involves substantial research efforts, such as complex electronics design, achieving IP68 waterproofing, lightweight structures, and battery safety, that could constitute a separate study. In essence, building physical and simulated AAMs presents a chicken-and-egg problem, where each development phase influences the other.

% While our research presents a robust simulation environment, we acknowledge the limitation of not being able to verify the Sim2Real transfer fully. Developing a fully functional physical AAM involves substantial research efforts that could constitute a separate study, encompassing challenges like electronics integration, on-board GPUs, waterproofing to at least IP68 standards, power supply, lightweight design, and battery safety. In essence, building physical and simulated AAMs presents a chicken-and-egg problem, where the development of physical and simulated AAMs influences each other. However, it's important to note that many highly regarded works that develop expensive photorealistic simulators, such as \cite{savva2019habitat,szot2021habitat,li2021igibson,gulino2024waymax,amer2023unav,puighabitat}, do not include physical robot experiments due to the significant challenges involved. Our AAM-SEALS distinguishes itself from these works by not only providing a state-of-the-art simulation environment but also introducing a novel class of robots with carefully designed morphology, kinematics, dynamics, and control systems. We believe AAM-SEALS opens a broad avenue for future research efforts:

Nonetheless, AAM-SEALS opens a broad avenue for future research efforts:

\begin{itemize}[left=0pt]
    \item \textbf{Supporting the Development of Physical AAMs:} Leveraging AAM-SEALS to aid in the development of physical AAMs is an ongoing project. This simulator will provide critical insights and validation before constructing physical prototypes.
    \item \textbf{Modeling Power Depletion Effects:} Future versions of AAM-SEALS could incorporate the impact of battery depletion on motors, control, and planning, as exemplified in \cite{bhatti2019genetically}. This addition would enhance the realism and practical utility of the simulations, allowing for more accurate modeling of how power limitations affect AAM performance.
    \item \textbf{Enhanced Manipulation:} Attaching a second manipulator to the AAM and exploring the resulting novel opportunities within AAM-SEALS could significantly expand the robot's capabilities.
    \item \textbf{Efficient Simulation:} Simulating particle physics is computationally expensive. Future research could focus on developing a hierarchical simulation approach, where high-resolution simulations are limited to the local region around the AAM. This localized high-resolution simulation would be informed by an outer, lower-resolution simulation that incorporates broader environmental factors such as temperature, depth, and ocean flow.
\end{itemize}

\section*{ACKNOWLEDGMENT}

We would like to thank Marcelo Jacinto, the author of the Pegasus simulator \cite{jacinto2023pegasus}, for his insightful debugging advice during a meeting with our team. Our appreciation also goes to Charlie Hanner for providing essential diving support for our real-world experiment, which measured acceleration changes over time as objects fell into water. Additionally, we are grateful to Isha Hemant Surjuse and Ian Fuller for their invaluable guidance and support in 3D-printing the Aerial-Aquatic Manipulator, which was instrumental to the success of this project. Finally, we thank Guanqun Luo for her invaluable contributions to the creation of Fig.~\ref{fig:intro}.

\bibliographystyle{plainnat}
\bibliography{paper}

\begin{appendices}



\onecolumn

% This appendix aims to address potential concerns and questions detailed in Appendix.~\ref{Ap:QA}, provide further details on position-based dynamics in Appendix.~\ref{Ap:PBD}, elaborate on AAM dynamics and control in Appendix.~\ref{Ap:AAM_Control}, and discuss additional experiments in Appendix.~\ref{Ap:RL}. 

% In our supplemental video, we showcase several aerial-aquatic trajectories used to complete various tasks. Due to hardware constraints and the high cost of running liquid simulations, reinforcement learning in aerial-aquatic environments is notably slow. Consequently, we will add more demonstration videos to the anonymous Google Drive: \url{https://drive.google.com/drive/folders/1ghE0sFIZ1hNrTdV5EBSUGiNVyWthGjrN?usp=sharing}

\section{Potential Questions and Our Answers} \label{Ap:QA}
In this section, we aim to address potential concerns and questions and
hope to clear any doubts or uncertainties that may arise.

\qund{You mentioned that your SEALS is based on Isaac Sim, while the AAM control is based on Pegasus. Can you specify your unique contribution?} 

To the best of our knowledge, there currently exists no high-fidelity simulator capable of effectively modeling movement both underwater and in the air. Although position-based dynamics \cite{macklin2014unified, muller2007position, macklin2013position} incorporated in the powerful NVIDIA Isaac Sim framework\footnote{https://developer.nvidia.com/isaac/sim} seems promising, their application in the development of a high-fidelity underwater robotics simulator in fluids of free space has not yet been explored. Additionally, adapting these dynamics to support motion across both aerial and aquatic mediums, including quadcopter dynamics, presents further challenges. Our initial attempt to create such a simulator required significant effort. 

To determine the suitability of leveraging position-based dynamics, one of our major tasks was to integrate traditional rigid-body-based hydrodynamics, as used in the cutting-edge photorealistic simulator UNav-Sim, into the Isaac Sim framework (specifically AAM-SEALS) alongside position-based hydrodynamics. This integration is non-trivial and allows us to compare the two hydrodynamics models, providing valuable insights.

The application of the control design from \cite{5980409} to AAM control is also not straightforward, as it is designed solely for aerial robotics to generate trajectories, without manipulators or underwater environments. Our AAM-SEALS adapted the control logic from Pegasus \cite{jacinto2023pegasus} by incorporating manipulator control. We specifically tailored our design for the morphology, kinematics, sensing, and control of our AAM, featuring a thinner arm, a three-finger gripper, and two RGB-Depth cameras (one looking ahead and one looking down) for aerial-aquatic manipulation tasks.

Our final contribution is a comprehensive evaluation of both the robot and the simulator, including applications in visual reinforcement learning (RL).

\qund{Is there a weakness in not including experiments on Physical AAMs?}

Not necessarily. Many advanced simulator studies (e.g., \cite{savva2019habitat,szot2021habitat,li2021igibson,gulino2024waymax,amer2023unav,puighabitat}) prioritize simulation development over physical experiments due to practical constraints. Building on these works, we validated our simulator’s hydrodynamic accuracy through real-world experiments with 3D-printed AAM models, demonstrating realistic acceleration patterns during air-to-water transitions. While these tests confirm hydrodynamic accuracy, they do not fully capture the complexity of cross-medium tasks. This underscores the role of AAM-SEALS as a crucial foundation for guiding future physical AAM development.

% \qund{How are the force term in Equ. 4 and another force term in Equ. 3 connected?}

% \qund{Evalaution with more RL models?}
% We have included those results in Appendix.~\ref{Ap:RL}

\qund{What's the action space and reward function for the RL?} %Did you use random seeds to train RL multiple times?}

Please refer to Appendix.~\ref{Ap:RL} for details.

% \qund{Would you consider other robot learning paradigms such as imitation learning?}

\qund{Could you explain more regarding position-based hydrodynamics?}

Please refer to the Appendix.~\ref{Ap:PBD}.

\qund{Since Aerial-Aquatic Manipulators (AAMs) represent a novel class of robots, could you elaborate on how future researchers might design and develop other forms of AAMs?}

We have outlined a general guideline in Appendix.~\ref{Ap:custom_aam} to assist future researchers in designing and developing their own AAMs.

\qund{It's great that you have evaluated Reinforcement Learning from Demonstrations techniques, such as Soft Actor-Critic from Demonstrations (SACfD). Did you also consider standard Reinforcement Learning methods, like Soft Actor-Critic (SAC)?}

\begin{figure}[t] \vspace*{-0.1cm}
  \begin{center}
    \includegraphics[width=0.8\linewidth]{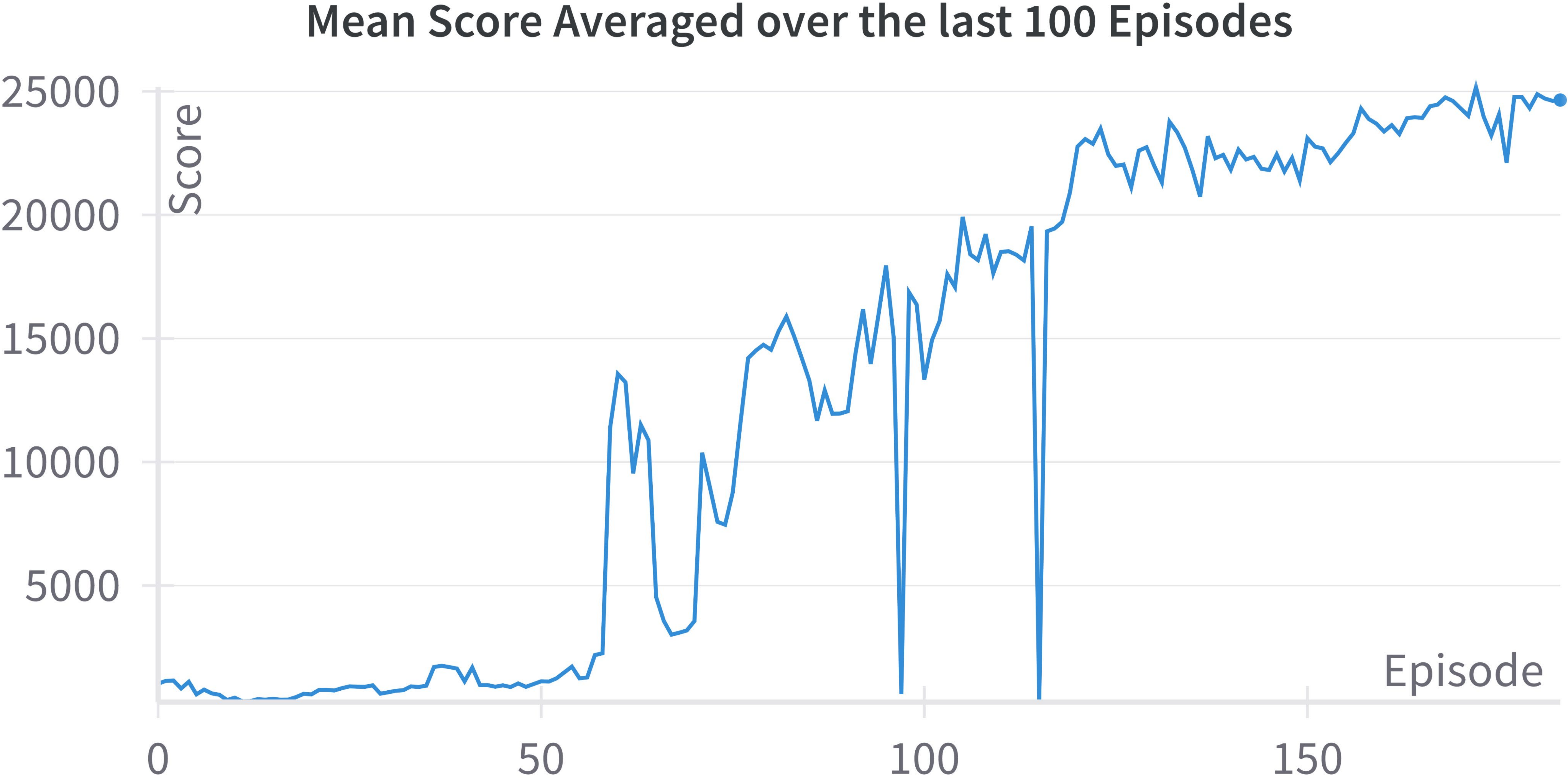}
  \end{center}  
  \caption{Reinforcement Learning (Soft-Actor Critic) results}
  \label{fig:RL_result}
\end{figure}

Yes, we have evaluated SAC as well. While it is effective, it performs slightly worse than SACfD -- it requires more training episodes to achieve meaningful performance improvements and ultimately converges to a lower score. Please refer to Fig.~\ref{fig:RL_result} for detailed results.

% \qund{Where can I find the description of manipulator modeling in Sec \ref{sssec:AAMDM} AAM Dynamics Modeling? Also, what does the expression $E_i += E_p$ mean in Sec.~\ref{sec:control_dev} Control Development?}

% Please refer to the Appendix.~\ref{Ap:AAM_Control} for details.

% \qund{How many trajectories did you collect for evaluating the position-tracking error?}% Can you justify why the amount is reasonable?}

% We collected 10 trajectories and believe this is sufficient to demonstrate the position-tracking error between dynamics systems. With 10 trajectories we could capture a fairly representative sample of directions, magnitudes, and sequences of actions. In addition, most of the trajectories were evaluated at more than 700 time points, meaning the 10 collected trajectories hold a large amount of evaluative data. The results of all 10 trajectories follow the trends presented in Sec.~\ref{section:Evaluation}.2 of the paper, as shown in Fig.~\ref{fig:dynamic_trajs}.

% \begin{figure}[H]
%   \centering
%   \includegraphics[width=1\textwidth]{figures/dynamics_trajectories.pdf}
%   \caption{Dynamics trajectory comparisons across all 10 collected trajectories}
%   \label{fig:dynamic_trajs}
% \end{figure}

\section{Preliminary Knowledge on Position-Based Dynamics} \label{Ap:PBD}

In the paper, the fluid simulation method uses the position-based dynamics (PBD) approach (\cite{macklin2014unified, muller2007position, macklin2013position}), which is closely related to the smoothed particle hydrodynamics (SPH) explained in \cite{monaghan1992smoothed,kanehira2022effects,fraga2019smoothed,zago2017simulating}. SPH is a well-known method that computes density and forces based on particle method for fluid simulation. However, SPH is sensitive to density fluctuations due to neighborhood deficiencies, and enforcing incompressibility is computationally expensive due to the unstructured nature of the model. SPH algorithms often become unstable if particles do not have enough neighbors for accurate density estimates. Typically, stability in SPH is maintained by taking sufficiently small time steps or using many particles, both of which increase computational costs.

In contrast, PBD improves upon these limitations by directly manipulating particle positions to satisfy physical constraints, specifically a density constraint given by:

\begin{equation} \label{eq:pbd_eq_0}
C(x_1,...,x_n) = \frac{\rho_i}{\rho_0} - 1 \leq  0
\end{equation}
where $\rho_i$ is the density at particle $i$ and $\rho_0$ is the rest density of the fluid. This ensures that particles maintain a proper distance from each other, effectively preventing clustering. PBD benefits from unconditionally stable time integration and robustness, making it popular with game developers and filmmakers. By addressing particle deficiencies at free surfaces and handling large density errors, PBD allows users to trade incompressibility for performance while remaining stable.

The PBD method also integrates additional effects such as cohesion and surface tension by adopting models such as those proposed by Akinci et al. (2013) \cite{akinci2013versatile}. For fluid-solid coupling, boundary particles are used to compute pressure forces between fluid and solid surfaces, ensuring accurate interactions. In the work \cite{macklin2013position}, the density estimation for fluid particles includes contributions from both fluid and solid particles, represented as:

\begin{equation} \label{eq:pbd_eq_1}
\rho_i = \sum_j m_j W(x_i - x_j,h) 
\end{equation}
where $m_j$ is the mass of particle $j$, $W$ is the smoothing kernel, $h$ is width of the smoothing kernel $W$, and $x_i - x_j$ is the distance between particles $i$ and $j$. This approach could then improved to a mass-weighted version of position-based dynamics as proposed in the unified position-based dynamics work \cite{macklin2014unified}:

\begin{equation} \label{eq:pbd_eq_2}
\rho_i = \sum_{fluid} m_j W(x_i - x_j,h) + s\sum_{solid} m_j W(x_i - x_j,h) 
\end{equation}
where a parameter $s$ is introduced to account for the differing densities, which allows for the realistic simulation of buoyancy and sinking behaviors of objects with different densities.

\section{Additional Reinforcement Learning Experiments}\label{Ap:RL}

In this section, we report the settings and results of reinforcement learning using pose states.

\mund{State Space}
Our state space includes the positions and orientations of both the object to be grasped and our AAM. The state space also includes the velocities of our AAM along the x, y, and z directions.

\mund{Action Space}
The action space in our framework is defined as a 3-tuple: [velocity\_x, velocity\_y, velocity\_z], representing the AAM's body movement with respect to the world frame. It is assumed that once the AAM's gripper reaches a desired position, there exists an engineered policy to automatically close the fingers, grasp the object, and then ascend out of the water.

\mund{Termination Condition}
% In our RL environments, the termination conditions are designed to determine whether an episode is a Success or a Failure. Success is achieved when the object is catched and lifted above the water. However, the RL environment can terminate with a Failure under two conditions:

In our RL environments, termination conditions are designed to determine the success or failure of an episode. Success is achieved when the AAM's gripper reaches within a distance of 0.01 meters of the target object.

\mund{Reward Function}

The reward function is defined on the basis of the distance between the AAM and the object, adjusted with a height offset to ensure clearance for grasping. It categorizes distances into three regions: outer (distance greater than 1 meter), inner (distance between 1 meter and $d_{t}$), and success (distance less than $d_t$). Each region employs a different reward calculation to provide dense rewards instead of sparse ones, with an added exponential growth factor to amplify the rewards as the AAM approaches the object. The reward formulations for each region are as follows:

\begin{itemize}
    \item \textbf{Outer Region:} $r = \exp(-d)$
    \item \textbf{Inner Region:} $r = \frac{1}{d}$
    \item \textbf{Success Region:} $r = 1000 \frac{1}{d_{t}}$
\end{itemize}

Here, $d$ denotes the Euclidean distance between the AAM and the object, and $d_{t}$ is the distance threshold below which the distance is considered a success. During our training, $d_{t}$ was set to $1 \times 10^{-2}$ meters.

To deter erratic behavior, penalty rewards are implemented for the RL agent. If the AAM's velocity surpasses a defined threshold, it incurs a penalty of -5. Furthermore, if the AAM achieves the success condition but subsequently leaves the success region in a specified number of steps, it receives a penalty equivalent to $-1000 \frac{1}{d_{t}}$. This penalty is intended to enforce stability and keep the AAM's gripper within the success region during operation.

% \mund{Hard-coded Policy for Fingers Closure and Lifting Up}
% Once inside the success region to grasp the object, the drone is given negative velocity in the z-direction to move towards the object. When the drone reaches a certain height threshold, the gripper is closed to grasp the object. After grasping, positive z-velocity is given to make the drone fly up towards the final destination, and a higher success reward for reinforcement learning is provided as $1000 \frac{1}{d_{t}}$.

% \mund{RL Agents}

\subsection{Reinforcement Learning from Demonstrations Hyperparameters}
In this section, we provide the values of the crucial hyperparameters listed in Table~\ref{tbl_result}.

% https://tex.stackexchange.com/questions/63204/trying-to-replicate-a-table-from-academic-paper
\begin{table}[h]
\begin{center} {\footnotesize
\begin{tabular}{cc}
\hline
 Parameter 
 & Value\\
\hline
batch size & 2048 \\[0ex]
initial random actions & 10000 \\[0ex]
total timesteps & 10,000,000 \\[0ex]
episode length & 1,000 \\[0ex]
distance threshold ($d_t$) & $10^{-2}$m \\[0ex]
Isaac Sim physics simulation timestep ($dt$) & $0.004$ \\[0ex]
replay buffer size & 100,000 \\[0ex]
learning rates for actor and critic & 0.006 \\[0ex]
discount ($\gamma$) & 0.99 \\[0ex]
exploration noise & 0.1 \\[0ex]
minimal exploration noise & 0 \\[0ex]
N-Step & 10 \\[0ex]
pretrain step & 5,000 \\[0ex]
PER\_Alpha & 0.6 \\[0ex]
PER\_Beta & 0.4 \\[0ex]
PER\_EPS & 1e-6 \\[0ex]
PER\_EPS\_DEMO & 1.0 \\[0ex]
number of hidden layers (all networks) & 2 \\[0ex] % 32.94
number of hidden units for layer 1 and 2 & [256, 256] \\[0ex] % 32.94
nonlinearity & ReLU \\[0ex] % 32.94
seeds & 0 \\[0ex]

% & $0.025^*$ & -0.002 & $1.155^*$ \\[-2ex]
\hline
\end{tabular} }
\end{center}
\caption{Reinforcement Learning Hyperparameters}
\label{tbl_result}
\end{table}

\yantian{
\section{Modeling, Control, and Learning of Aquatic Animals} \label{Ap:Aquatic_Animals}

The crab mesh was created in Blender 3.3, where we used various mesh tools to sculpt the body, legs, and claws of the crab into a detailed and realistic model. Once the basic structure was complete, we activated the Phobos extension \cite{phobos}, a powerful tool for robotics modeling. With Phobos, we assigned joints to key parts of the crab, such as the leg bases and claw hinges, by selecting the corresponding mesh segments and establishing them as joints. We then linked these joints to simulate natural movements, ensuring that each leg and claw could articulate correctly. Fig.~\ref{fig:crab_mesh} Careful naming and organization of components within Phobos allowed for a clean and manageable hierarchy, which is essential for future robotics applications. Finally, we exported the entire setup, including joints and links, in URDF format using the Phobos export function, making the crab model ready for integration into the robotics simulation environment. To ensure everything worked as expected, we loaded the URDF into an Isaac Sim and fine-tuned the model, checking that the crab's movements were accurately represented and adjusting any discrepancies.

\begin{figure}[H]
  \centering
  \includegraphics[width=1\textwidth]{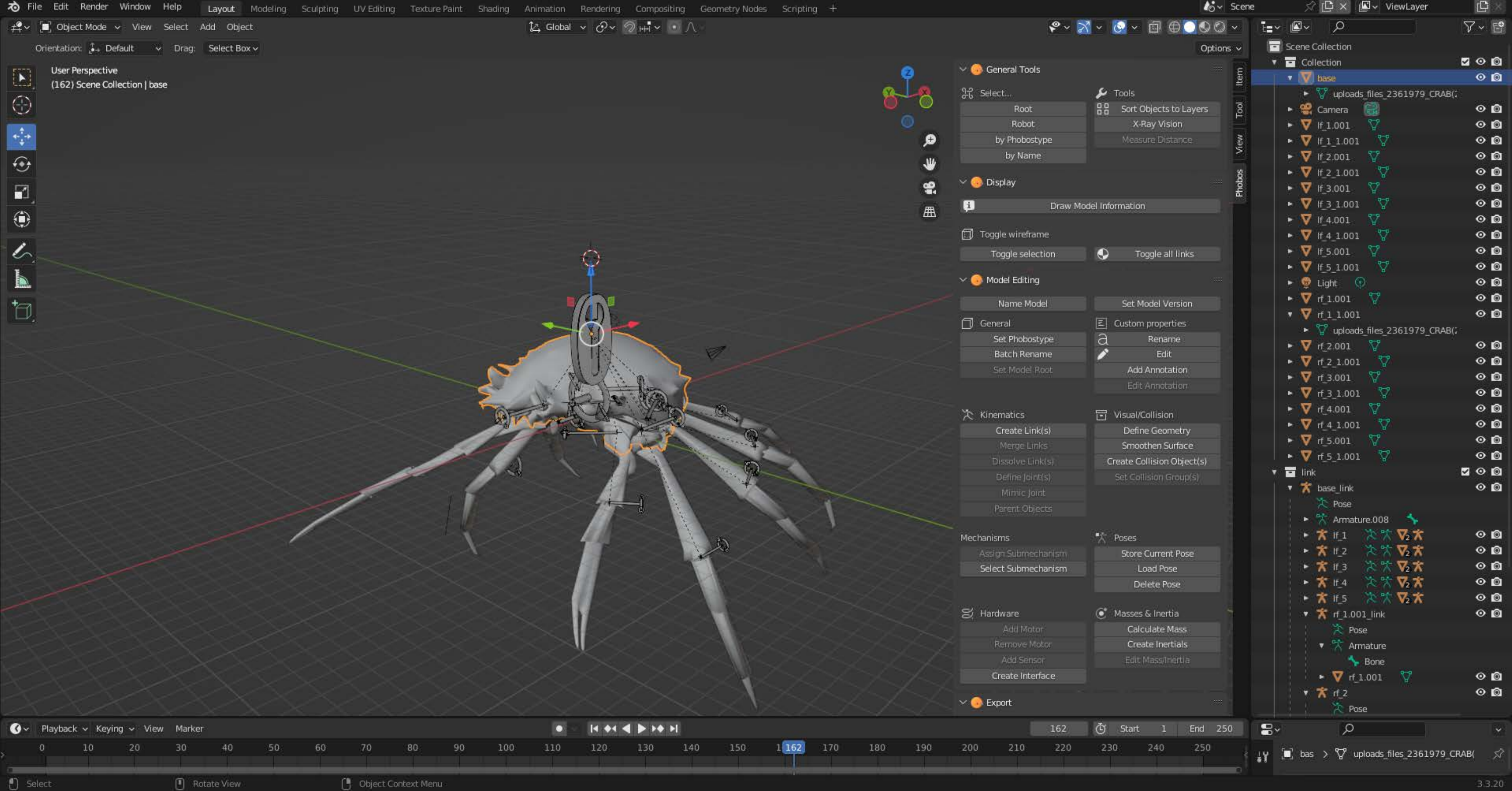}
  \caption{Crab Mesh in Blender}
  \label{fig:crab_mesh}
\end{figure}

After creating the URDF and meshes of the crab, we then need to model it and assign controllers to joints. The crab model consists of 18 joints, resulting in a total of 18 degrees of freedom (DoF) for the body. Managing such a high-DoF agent can be challenging, so we employed a reinforcement learning (RL) based policy for control. The controller used for each joint is a position-based controller, defined by the following equation:

\[
F = K_d(V_d - \text{current\_V}) + K_s(P_d - \text{current\_P})
\]

Where:
\begin{itemize}
    \item $K_d$ is the damping coefficient,
    \item $K_s$ is the stiffness coefficient,
    \item $V_d$ is the desired velocity (typically set to 0),
    \item $\text{current\_V}$ is the current velocity of the joint motion,
    \item $P_d$ is the desired position (angular position in radians),
    \item $\text{current\_P}$ is the current angular position of the joint.
\end{itemize}

The reward function used in the RL approach is similar to the one described in Appendix \ref{Ap:RL}. The objective is for the robot to reach a fixed target position, with the robot spawning from different starting positions each time. That said, future works could consider more advanced learning algorithms such as goal-conditioned reinforcement learning and even adversarial multi-agent reinforcement learning that can empower the crab with defensive strategies.

\section{Adding your Customized AAM}\label{Ap:custom_aam}
In this paper, we provide a high-level description how future researchers could create their own AAM and load into our SEALS. We will release a detailed tutorial online once the paper gets accepted.

To include a different robot model in our simulator, you will need a .usd (Universal Scene Description) file of the robot. The process involves several steps:

\begin{enumerate}
   \item Designing the Robot Model:
   \begin{itemize}
     \item Begin by designing the robot model in SolidWorks (a 3D CAD Design Software)
     \item Create the individual parts and assemble them, ensuring all joints and kinematic properties are accurately defined
   \end{itemize}
   \item Generating the Mesh Files and .urdf File:   \begin{itemize}
     \item Create mesh files to represent the robot's physical structure visually and geometrically
     \begin{description}
     \item[Note:] These meshes provide a realistic appearance in the simulation and can be exported alongside the .urdf file
     \end{description}
     \item Export these meshes alongside the .urdf (Unified Robot Description Format) file
     \begin{description}
     \item[Note:] The .urdf file encapsulates the working joints, linkages, and their respective constraints
     \end{description}
   \end{itemize}

   \item Importing into Isaac Sim:
   \begin{itemize}
   \item Import the .urdf file into the Isaac Sim simulator
   \item Convert the .urdf file into a .usd (Universal Scene Description) file
   \begin{description}
     \item[Note:] The .usd format is essential because it enables seamless integration and manipulation within the simulator, ensuring that all joints operate correctly and the robot's physical characteristics are preserved
     \end{description}
   \end{itemize}
\end{enumerate}

}
\end{appendices}

\end{document}